\documentclass[11pt]{article}
\usepackage[T1]{fontenc}
\usepackage[utf8]{inputenc}
\usepackage{lmodern}
\usepackage[margin=1in]{geometry}
\usepackage{amsmath,amssymb}
\usepackage{graphicx}
\usepackage{booktabs}
\usepackage{multirow}
\usepackage{array}
\usepackage[table]{xcolor}
\usepackage{enumitem}
\usepackage[font=small,labelfont=bf]{caption}
\usepackage[numbers,sort&compress]{natbib}
\usepackage{microtype}
\usepackage{xurl}
\usepackage[colorlinks=true,linkcolor=blue!55!black,citecolor=blue!55!black,urlcolor=blue!55!black]{hyperref}

\newcommand{\mtus}{\textunderscore\allowbreak}
\newcommand{\mt}[1]{{\ttfamily\let\_\mtus#1}}  
\newcommand{\pair}[2]{#1\,/\,#2}          
\newcommand{\mxf}{MXFP8}
\newcommand{\tw}{FP8-TW}                   
\newcommand{\bw}{FP8-BW}                   
\newcommand{\ms}{\,$\mu$s}
\setlist{itemsep=2pt,topsep=3pt}

\title{End-to-End MXFP8 Reinforcement Learning on Blackwell:\\What Matching Buys, When It Pays Off, and How It Fails}
\author{Wengee Zhang\\[2pt] {\small\texttt{wengeezhang@gmail.com}}}
\date{}

\begin{document}
\maketitle

\begin{abstract}
Low-precision arithmetic can make reinforcement learning (RL) for large language models cheaper, but RL requires the
sampler and the learner to agree. Several frameworks now run FP8 on both sides. We
implement end-to-end MXFP8 RL in verl---Megatron/Transformer Engine training; vLLM and SGLang rollout; dense and MoE
models---and measure on B200 GPUs, with Qwen3-8B and Qwen3-30B-A3B, what this design buys, when it is faster, and how
it fails.
(1)~\emph{Matching.} For an MXFP8 learner, a sampler on the learner's grid is more consistent than a
higher-precision BF16 sampler: in one engine at fixed weights, train--rollout KL is 11\% lower with half of the
sampler's linear layers on the grid and 16\% lower with all of them, and the synchronized weights we checked equal the
learner's FP8 forward weights byte for byte. On the MoE model, an FP8 pipeline whose sampler uses another grid has twice the
shared-grid pipeline's KL from the first step. Without a quantized KV cache, BF16 end to end remains 2.5--10$\times$
more consistent.
(2)~\emph{Speed.} MXFP8 training adds a fixed cost per micro-batch and saves a per-token cost, so in our dense setting
it pays off only above about 4.5k tokens per micro-batch; a micro-batch sweep on identical data confirms the
per-micro-batch cost, and a trace shows the GPU waiting for the host for most of the added time. End to
end, MXFP8 cuts MoE step time by 10.7\% and generation time by 24\%, and slows a short-sequence dense workload by 5\%.
(3)~\emph{Correctness.} We state seven conditions every weight synchronization must preserve and checks that test them
against the live engine. Each of the seven failures we met on hardware broke one condition, five silently; the checks
caught two at their first opportunity, each missed by the other checks, and CPU fault-injection tests reproduce five.
All runs are single-seed and at most 60 steps long; we make no claim about final model quality.
\end{abstract}

\section{Introduction}
\label{sec:intro}

Reinforcement learning with verifiable rewards has become a standard stage of LLM post-training
\citep{shao2024deepseekmath,yu2025dapo}. An RL step has two GEMM-heavy halves: an inference engine such as
vLLM~\citep{kwon2023vllm} or SGLang~\citep{zheng2024sglang} \emph{samples} responses, and a distributed trainer such
as Megatron-Core~\citep{shoeybi2019megatron} \emph{learns} from them. Both halves are candidates for FP8 arithmetic,
which doubles peak tensor-core throughput over BF16 on Hopper and Blackwell GPUs. For pre-training, FP8 has been
shown to match BF16 at scale with per-tensor, block-wise, and microscaling recipes
\citep{micikevicius2022fp8,peng2023fp8lm,deepseekv3,mishra2025mxfp8}.

RL, however, has a requirement that pre-training lacks: the sampler and the learner must represent the \emph{same}
policy. In practice they are two separate implementations, and even in BF16 their log-probabilities differ; this
train--inference mismatch makes nominally on-policy training slightly off-policy, biases the policy gradient, and
has been linked to training collapse, which is why importance-sampling corrections are now common
\citep{yao2025offpolicy,liu2025speedkills,qi2025fp16}. Quantizing only one side widens the gap. A BF16 learner with an
FP8 or INT8 sampler \citep{liu2025flashrl,qiu2026fp8rl} speeds up rollout, but the sampler then serves a
post-training-quantized copy of the policy; an FP8 learner with a BF16 sampler speeds up training, and the sampler
then serves a policy the learner never computes with.

The field has therefore moved to low precision on \emph{both} sides. On Hopper GPUs, unified block-wise FP8
pipelines keep the learner and the sampler on one FP8 recipe \citep{jetrl2026,miles2025unifiedfp8,nemorl2026fp8}. On
Blackwell (SM100), whose tensor cores implement the OCP microscaling (MX) formats~\citep{ocp2023mx,rouhani2023mx}
natively---MXFP8 stores E4M3 elements in blocks of 32 that share one power-of-two (E8M0) scale, applied inside the
matrix-multiply instruction---Miles runs MXFP8 end to end with a byte-exact quantizer contract between Transformer
Engine and FlashInfer \citep{miles2026mxfp8}, and SkyRL makes the synchronized FP8 weights bitwise identical to the
trainer's after measuring a mean relative error of 0.038 between the two sides' FP8 weights when the sampler
re-quantized independently \citep{skyrl2026fp8}. The reverse is also documented: a 688B-parameter mixture-of-experts (MoE) model whose MXFP8 learner was paired with a
block-wise FP8 sampler collapsed, and truncated importance sampling did not prevent it \citep{skt2026axk2}. We call the
arrangement in which the sampler receives the learner's own quantized weight grid \emph{shared-grid} weight
synchronization. It needs no calibration and no offline checkpoint conversion: the sampler starts from the BF16
checkpoint in low-precision mode and receives freshly quantized weights after every optimizer step.

These reports establish that shared-grid synchronization is feasible and that mismatched grids can be harmful. They
leave three practical questions open, which this paper addresses with measurements on B200 GPUs:

\begin{itemize}
\item \textbf{Q1 (consistency).} How much does matching buy, and relative to what? In particular, for a
low-precision learner, is a matched low-precision sampler more or less consistent than a BF16 sampler, and does the
answer depend on how much of the sampler is matched?
\item \textbf{Q2 (speed).} When does end-to-end MXFP8 make an RL step faster, and why does it sometimes make it
slower? SkyRL observed that per-micro-batch host overhead made its FP8 training phase roughly time-neutral on H100
\citep{skyrl2026fp8}; we turn that observation into a threshold, test it on a held-out workload and in a controlled
sweep, and profile the cost.
\item \textbf{Q3 (correctness).} What must a weight synchronization preserve for the sampler to keep computing the
learner's policy, and how can violations that do not crash be detected at run time?
\end{itemize}

\paragraph{Contributions.}
\begin{enumerate}
\item \textbf{Consistency measurements} (Section~\ref{sec:consistency}) in four settings, anchored at the first step,
where all arms of a setting hold the same weights and see the same prompts, and a controlled experiment in one engine
at fixed weights. The controlled experiment orders samplers by how many of their linear layers are on the learner's
grid (none, half, all) with confidence intervals from prompts paired across arms, and shows that for an MXFP8 learner the
matched sampler is more consistent than a BF16 one; a byte-level comparison confirms that the synchronized weights
equal the learner's FP8 forward weights. The settings add an FP8-blockwise pipeline whose sampler grid differs from
its learner's on Blackwell, and the drift of per-tensor FP8 over training, which accounts for most of the
late-training differences between recipes.
\item \textbf{An empirical break-even model for low-precision RL training} (Section~\ref{sec:speed}). We fit each
stage's time as a fixed cost plus a per-token cost and find that the added fixed cost is a cost per micro-batch, so the
threshold is a number of tokens per micro-batch, $m^\ast=\Delta\delta/\Delta c$. The algebra is elementary; the
contribution is the measurement. Fitted on one workload, the model explains why that workload measured no speed-up,
predicts a held-out long-sequence workload better than a per-step model, and is consistent with the same recipe
becoming a net loss on a slower host. A sweep of micro-batch sizes on replayed, identical data confirms the
per-micro-batch form and crosses zero near the fitted break-even point, without pinning that point down by itself,
and a trace of the added time shows the GPU idle for most of it, waiting for the host to issue the next operation.
\item \textbf{A correctness specification for the train--inference boundary} (Section~\ref{sec:failures}). We
state seven conditions that every weight synchronization must preserve. They cover the layer set, the weight grid,
kernel state derived at load, CUDA-graph storage, the completeness of a streamed update, the sampler's output, and
the checks' own observability. For three run-time checks we state which conditions each tests and what it cannot
see. Each of the seven failures we met on B200 broke one condition, five of them without an error. On real faults
the checks are complementary: the audit alone caught F2, and the self-check alone caught F4. Five of the failures are
reproduced by CPU fault-injection tests. A hardware fault-injection protocol specifies how to measure the checks'
detection rates, false alarms, delay and cost.
\end{enumerate}

\paragraph{Scope.} All runs use one seed per configuration. The runs of the four settings last 10--60 RL steps at
learning rate $10^{-6}$; the 17 controlled runs (Section~\ref{sec:setup}) last 3--12 steps, at learning rate 0 or
$10^{-6}$. They measure train--rollout consistency, per-stage speed and memory, and correctness checks; they are too short to rank final
model quality, and we make no such claim (Section~\ref{sec:limitations}).

\section{Background}
\label{sec:background}

\subsection{FP8 formats and scaling granularity}
FP8 training stores tensors in E4M3 (4 exponent, 3 mantissa bits, max 448) or E5M2 and rescales each tensor, or
part of a tensor, into the representable range~\citep{micikevicius2022fp8}. Scaling granularity is the main design
axis. \emph{Per-tensor} scaling uses one FP32 scale per tensor, computed from the current absolute maximum
(``current scaling'') or from a history of maxima (``delayed scaling''). \emph{Block-wise} scaling, as used for
DeepSeek-V3, uses one FP32 scale per $1\times128$ tile of activations and per $128\times128$ tile of
weights~\citep{deepseekv3}. \emph{Microscaling} (MX) uses one shared scale per 32 consecutive elements along the
reduction dimension, and the scale is an 8-bit exponent (E8M0), i.e.\ a power of two~\citep{ocp2023mx,rouhani2023mx};
MXFP8 therefore costs 8.25 bits per element. A power-of-two scale is exact to apply, but when it is rounded so that no
element overflows, the block maximum can land up to a factor of two below the format's maximum, leaving up to one
binade of the element range unused. E4M3 spans about 18 binades, so the loss is small at 8 bits; a 4-bit E2M1 element
spans fewer than four, which is one reason NVFP4 uses smaller blocks with non-power-of-two
scales~\citep{nvidia2025nvfp4}.

\subsection{FP8 training in Transformer Engine, and what changes on Blackwell}
Transformer Engine (TE)~\citep{te} keeps high-precision master weights and quantizes the operands of every linear
layer on the fly: the input and weight for the forward GEMM (fprop), the output gradient and the weight for the
data-gradient GEMM (dgrad), and the input and output gradient for the weight-gradient GEMM (wgrad). TE exposes the
three granularities above as the recipes \mt{Float8CurrentScaling} (tensorwise), \mt{Float8BlockScaling}
(blockwise) and \mt{MXFP8BlockScaling}. Because wgrad reduces over tokens, MXFP8 quantizes some operands in
32-element blocks along the \emph{token} dimension, so every dimension of a quantized tensor must be a multiple
of~32, including the packed sequence dimension.

Blackwell's block-scaled MMA instructions take MXFP8 operands together with their E8M0 scales, so the tensor cores
apply the scales. Hopper's FP8 MMA takes no scales; block-wise recipes there apply FP32 block scales outside the MMA
instruction, inside library GEMM kernels. TE~2.16 uses such a cuBLASLt GEMM for \mt{Float8BlockScaling} on Hopper
only; on SM100 it \emph{emulates} the recipe with MXFP8 GEMMs by converting each FP32 block scale to E8M0, a
conversion that keeps only the exponent bits. TE therefore requires power-of-two block scales on SM100 and raises an
assertion otherwise (Section~\ref{sec:failures}, F6). Accumulation precision also
differs by generation: on Hopper, FP8 MMA accumulates with roughly 13--14 fractional bits, which is why DeepSeek-V3
promotes partial sums to FP32 CUDA-core registers every 128 elements~\citep{deepseekv3,khattak2025accum}; measurements
of Blackwell's non-block-scaled FP8 MMA report 25 fractional bits~\citep{khattak2025accum}. We are not aware of a
published measurement for the block-scaled instruction that MXFP8 uses, and we did not make one.

\subsection{The RL loop and train--rollout mismatch}
In each step the sampler generates responses and records the log-probability $\log\pi_{\text{roll}}(y_t)$ of every
sampled token; the learner recomputes $\log\pi_{\text{train}}(y_t)$ (the ``old log-prob'' pass), optionally computes
a reference-model log-prob for a KL penalty, and then runs the policy update; finally the new weights are sent to
the sampler. We use the mismatch diagnostics that verl~\citep{sheng2024hybridflow} logs on every step, all computed over
sampled response tokens:
\begin{itemize}
\item \mt{rollout\_corr/kl}: the mean over sampled tokens of
$\log\pi_{\text{roll}}(y_t)-\log\pi_{\text{train}}(y_t)$, a direct estimator of
$\mathrm{KL}(\pi_{\text{roll}}\,\|\,\pi_{\text{train}})$ under the sampler's distribution;
\item \mt{rollout\_corr/chi2\_token}: $\mathbb{E}[\rho_t^2]-1$ with $\rho_t=\pi_{\text{train}}(y_t)/\pi_{\text{roll}}(y_t)$,
which equals the variance of the token-level importance weight because $\mathbb{E}_{\pi_{\text{roll}}}[\rho_t]=1$;
\item \mt{rollout\_corr/log\_ppl\_abs\_diff}: the mean over sequences of the absolute difference in per-token
log-perplexity;
\item the Pearson correlation of the two sampled-token probabilities~\citep{minimax2025m1}, and the mean absolute
probability difference.
\end{itemize}
These diagnostics are related: \mt{kl} and \mt{log\_ppl\_abs\_diff} are built from the same log-ratios, and the
Pearson correlation, \mt{chi2\_token} and the probability difference from the same probabilities. We therefore treat
agreement among them as consistency, not as independent confirmation.

\section{Learner and sampler precision: a taxonomy}
\label{sec:taxonomy}

Table~\ref{tab:taxonomy} organizes RL pipelines by the precision of the learner's GEMMs, the precision of the
sampler's GEMMs, and whether the sampler's quantized weights are the ones the learner computes with. The third column
is the axis this paper measures. Two quantized pipelines can use the same format and still sit on different grids:
on Blackwell, verl's FP8 end-to-end recipe trains with power-of-two block scales (required there) while its weight
synchronization quantizes with FP32 scales $\mathrm{absmax}/448$, so the two sides round the same BF16 weight
differently.

\begin{table}[t]
\centering
\small
\caption{Learner/sampler precision pairings. ``Grid'' is the relation between the sampler's weights and the
weight operand of the learner's forward GEMM. Rows marked $\checkmark$ are measured in this paper.}
\label{tab:taxonomy}
\resizebox{\linewidth}{!}{%
\begin{tabular}{@{}llllc@{}}
\toprule
Learner GEMMs & Sampler GEMMs & Grid relation & Reported in & Here \\
\midrule
BF16 & BF16 & same values, different kernels & standard RL; \citep{qi2025fp16,he2025nondeterminism} & $\checkmark$ \\
BF16 & FP8 / INT8 & post-training quantization of the learner & \citep{liu2025flashrl,qiu2026fp8rl,jetrl2026} & \\
FP8 tensorwise & BF16 & sampler not quantized & this work & $\checkmark$ \\
MXFP8 & BF16 & sampler not quantized & this work (ablation) & $\checkmark$ \\
MXFP8 & FP8 blockwise & different formats & \citep{skt2026axk2} (reward collapse) & \\
FP8 blockwise & FP8 blockwise & different scales (SM100: power-of-two vs.\ FP32) & this work & $\checkmark$ \\
FP8 blockwise & FP8 blockwise & shared recipe, Hopper & \citep{jetrl2026,miles2025unifiedfp8,nemorl2026fp8} & \\
FP8 / MXFP8 & FP8 / MXFP8 & shared, bitwise identical or byte-exact & \citep{skyrl2026fp8,miles2026mxfp8} & \\
MXFP8 & MXFP8 & shared, re-quantized every step with the learner's quantizer & this work & $\checkmark$ \\
\bottomrule
\end{tabular}}
\end{table}

\section{System design}
\label{sec:system}

\begin{figure}[t]
\centering
\includegraphics[width=\linewidth]{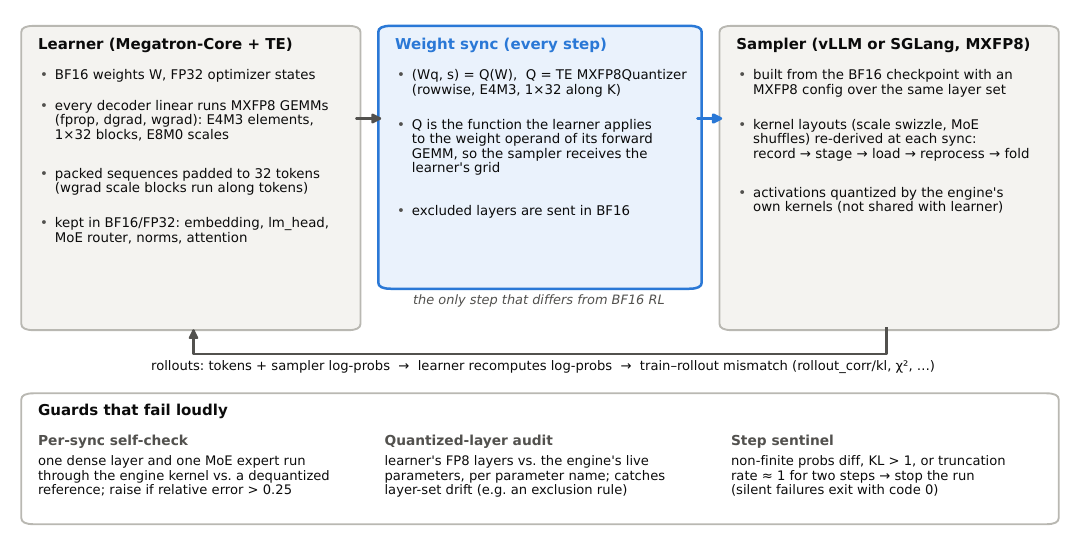}
\caption{End-to-end MXFP8 RL. The learner computes with MXFP8 GEMMs over BF16 master weights. At every
synchronization the weights are quantized with the learner's own TE quantizer, so the sampler receives the grid of
the learner's forward weight operand. The sampler re-derives its kernel layouts from those weights in place. Three
guards turn silent failures into errors.}
\label{fig:system}
\end{figure}

Figure~\ref{fig:system} shows the pipeline. Relative to BF16 RL, one step changes: the weight synchronization. We
describe the learner, the synchronization, the sampler integration, the policy for which layers stay in high
precision, and the guards.

\subsection{Learner}
The learner is Megatron-Core with Transformer Engine, configured through verl's existing override of the
Transformer Engine configuration (\mt{fp8="e4m3"} and \mt{fp8\_recipe="mxfp8"} under
\mt{override\_transformer\_config}). Weights and optimizer states stay in
BF16/FP32, so checkpoints are unchanged. The one functional gap was sequence padding: verl packs variable-length
sequences into a THD layout whose token dimension was aligned to $\mathrm{lcm}(16,\cdot)$ for the blockwise recipe,
while MXFP8's wgrad blocks run along tokens and need multiples of 32. Under the MXFP8 recipe the per-sequence
alignment becomes $\mathrm{lcm}(32,\cdot)$ (the mapping Megatron-Core applies to its own MoE FP8 padding); every
other recipe keeps its previous alignment. For MoE models we additionally enable Megatron's router padding for FP8
(\mt{moe\_router\_padding\_for\_fp8}). Attention, normalization, embedding, the output projection (\mt{lm\_head}) and
the MoE router are not TE FP8 linears and stay in BF16 or FP32.

\subsection{Shared-grid weight synchronization}
At each synchronization every weight of a quantized layer is passed through
\[
  (W_q, s) = Q(W), \qquad Q = \text{TE \mt{MXFP8Quantizer}}(\text{E4M3, rowwise, } 1\times32 \text{ blocks along } K),
\]
with rows padded to a multiple of 32 and the scales emitted as compact unswizzled UE8M0 bytes of shape
$[N, K/32]$. $Q$ is the quantizer TE applies to the weight operand of the learner's forward GEMM, so the sampler
receives the learner's forward weight grid; an independently written quantization kernel could round E8M0 scales
differently at block boundaries. For vLLM the quantization runs inside the vLLM worker process, which receives BF16
weights through verl's worker extension; for SGLang it runs on the trainer side before the weights are streamed.
Two properties are \emph{not} shared by construction and we do not claim them: the sampler quantizes activations
with its own kernels, and its GEMMs accumulate in a different order. Section~\ref{sec:bytes} compares the
synchronized bytes with the learner's cached FP8 weights.

\subsection{Sampler integration and the refit cycle}
\paragraph{Launch.} The sampler starts from the BF16 checkpoint with an injected quantization configuration: for
vLLM, a configuration that vLLM normalizes to its native ModelOpt MXFP8 method (E4M3 weights, UE8M0
\mt{weight\_scale}, block $[1,32]$); for SGLang, its FP8 method in MXFP8 mode. The configuration lists the layers to
keep in BF16 (Section~\ref{sec:layerset}).

\paragraph{Derived layouts.} After loading, serving kernels rewrite the weights and scales into kernel-specific
layouts: FlashInfer swizzles the scales, TRT-LLM's MoE kernels shuffle and interleave the expert weights, Marlin
repacks them. A weight update that writes the canonical tensors without redoing this work pairs fresh weights with
stale layouts. verl's FP8 path already handled this with a record/stage/reprocess cycle; we register the MXFP8
dense and MoE methods with it. At load, the cycle records each parameter's checkpoint layout; before a
synchronization it stages the parameter back to that layout; after loading it re-runs the method's
post-processing; and it folds the result into the original storage, so addresses captured by CUDA graphs stay
valid. Two MoE-specific repairs were needed. TRT-LLM's MXFP8 MoE preparation rewrites expert weights in place
without changing their shape or dtype, so the cycle's shape-based detection missed it; we now record same-shape
repacks explicitly (upstreamed as verl~\#7986). And vLLM's MoE method sets a ``post-processing done'' flag that made
later re-processing a no-op; the reprocess step clears it. For SGLang we add a custom weight loader, selected per
update request, that re-runs the post-processing of every MXFP8 layer whose resolved backend keeps derived copies.

\subsection{Which layers stay in high precision}
\label{sec:layerset}
The token embedding, \mt{lm\_head} and MoE routers stay in BF16 on both sides. \mt{lm\_head} is load-bearing: when it
was quantized, the logits became NaN and sampling degenerated while the run exited normally, and excluding it removed
the NaNs (F1 in Table~\ref{tab:failures}); we did not isolate the mechanism. The router is quantizable in vLLM's MoE blocks (it inherits the model's quantization
configuration), so it is excluded explicitly; SGLang's MoE blocks build it without a quantization configuration. The
exclusion list itself turned out to be a source of error: vLLM~0.24's ModelOpt matcher falls back to substring
matching, so the router entry \mt{model.layers.$i$.mlp.gate} also excluded the dense MLP's \mt{gate\_up\_proj} (F2).
The released code matches exclusions on module-path boundaries; the dense vLLM experiments in
Section~\ref{sec:setup} predate the fix.

\subsection{Guards}
Each guard targets a failure that otherwise produces plausible-looking metrics and exit code~0.
\begin{itemize}
\item \textbf{Per-synchronization self-check.} After the sampler re-processes its weights, the smallest MXFP8 dense
layer runs its own kernel on a random input and is compared with a BF16 GEMM on the dequantized canonical weights;
for MoE models one local expert is probed the same way by routing every probe token to it and comparing against a
dequantized gated-MLP reference. A relative error above 0.25 raises. Healthy layers show about 0.027 (dense) and
0.049 (MoE expert), the error of the kernel's activation quantization; stale layouts give $O(1)$.
\item \textbf{Quantized-layer audit.} At the first synchronization after a training step, the audit compares the set of
decoder layers that ran FP8 GEMMs in the learner with the sampler's rule and, on vLLM, with the engine's live
parameters, per parameter name.
\item \textbf{Step sentinel.} When the rollout is quantized, the trainer stops if the probability difference is
non-finite, if \mt{rollout\_corr/kl} exceeds 1.0, or if nearly every response hits the length cap on two
consecutive steps.
\end{itemize}
Section~\ref{sec:failures} states the conditions these guards test, what each can and cannot see, how they performed
on the faults we met, and how they are tested with injected faults.

\section{Experimental setup}
\label{sec:setup}

\paragraph{Hardware and software.} NVIDIA B200 GPUs (SM100); CUDA~13.0, PyTorch~2.11,
Transformer Engine~2.16.1, Megatron-Core~0.18, vLLM~0.24.0 (FlashInfer~0.6.12) and SGLang~0.5.12. Two host types
were used; because the host CPU turned out to set part of the step time (Section~\ref{sec:speed}), timings are
compared only between runs on the same host.

\paragraph{Settings.} Table~\ref{tab:settings} lists the four settings, which use two models of the Qwen3
family~\citep{qwen3}. All use GRPO with 64 prompts $\times$ 8
samples per step, mini-batch 32 prompts, learning rate $10^{-6}$, a KL loss to the reference model with coefficient
0.001 (\mt{low\_var\_kl}), no entropy bonus, and the same seed and data order in every arm of a setting. DAPO-Math-17k
is used as data with standard GRPO; we do not use the DAPO algorithm's dynamic sampling, clip-higher or overlong
shaping. Rollout-correction diagnostics are logged but no importance-sampling correction is applied.

\begin{table}[t]
\centering
\footnotesize
\caption{Experimental settings. A, B and D ran all their arms on one fast host (AMD EPYC~9555); C and the
\pair{\mxf}{BF16} arm of A ran on a host with a slower CPU (Intel Xeon~6960P), so their timings are not compared with
the others.}
\label{tab:settings}
\setlength{\tabcolsep}{4pt}
\begin{tabular}{@{}>{\raggedright\arraybackslash}p{2.1cm}>{\raggedright\arraybackslash}p{3.2cm}>{\raggedright\arraybackslash}p{3.3cm}>{\raggedright\arraybackslash}p{3.0cm}>{\raggedright\arraybackslash}p{3.6cm}@{}}
\toprule
& \textbf{A}: dense, gsm8k & \textbf{B}: dense, DAPO & \textbf{C}: dense, gsm8k, SGLang & \textbf{D}: MoE, DAPO \\
\midrule
Model & Qwen3-8B & Qwen3-8B & Qwen3-8B & Qwen3-30B-A3B (48 layers, 128 experts, top-8) \\
GPUs, parallelism & 2 B200, TP 2 & 2 B200, TP 2 & 2 B200, TP 2 & 4 B200, TP 2 $\times$ EP 2 \\
Prompt / response cap & 1024 / 2048 & 1024 / 8192 & 1024 / 2048 & 1024 / 8192 \\
Tokens per step & 0.29--0.68\,M & 2.9--3.7\,M & 0.37--0.67\,M & 3.4\,M \\
Steps & 60 & 32 & 30 & 10 \\
Micro-batching & 4 sequences (update), 8 (log-prob) & dynamic, 12\,288 tokens/GPU & as A & dynamic, 20\,480 tokens/GPU \\
Other & -- & fp8 KV cache (all arms) & sampler is SGLang & parameter offload, optimizer on CPU, full recompute, FlashInfer autotuning off \\
Validation & gsm8k test set, steps 0 and 60 & none & steps 0 and 30 & none \\
Arms & \pair{BF16}{BF16}, \pair{\tw}{BF16}, \pair{\mxf}{BF16}, \pair{\mxf}{\mxf} & \pair{BF16}{BF16}, \pair{\tw}{BF16}, \pair{\mxf}{\mxf} & \pair{BF16}{BF16}, \pair{\tw}{BF16}, \pair{\mxf}{\mxf} & \pair{BF16}{BF16}, \pair{\tw}{BF16}, \pair{\bw}{\bw}, \pair{\mxf}{\mxf} \\
\bottomrule
\end{tabular}
\end{table}

\paragraph{Arms.} ``\pair{X}{Y}'' denotes learner precision X and sampler precision Y. \tw{} is TE's per-tensor
current scaling in E4M3 with a BF16 sampler. \pair{\mxf}{BF16} is the \mxf{} learner with an unquantized sampler, an
ablation of the matched sampler. \pair{\bw}{\bw} is verl's FP8 end-to-end recipe (blockwise training, blockwise FP8
rollout) with the E4M3 format and power-of-two block scales, which Blackwell requires. The reference model is BF16 in
every arm. All learners use FP32 accumulation in the optimizer. In setting D, for memory reasons, the parameters are
offloaded to the host between stages and the optimizer runs on the CPU, and FlashInfer's kernel autotuning is off;
these settings are identical in all four arms.

\paragraph{The dense vLLM samplers were partially matched.} In settings A and B the vLLM sampler kept every
\mt{gate\_up\_proj} in BF16 because of the exclusion-matching bug described above (F2), while \mt{qkv\_proj},
\mt{o\_proj} and \mt{down\_proj} were on the learner's grid. For Qwen3-8B, \mt{gate\_up\_proj} holds 52\% of each
decoder layer's linear weights. The SGLang sampler (setting C) matches on module-path boundaries and had every decoder
linear on the grid, as did the MoE sampler (setting D), whose layers contain no dense \mt{gate\_up\_proj}; in D the
audit confirmed that learner and engine quantize the same 48 of 48 layers (18\,867 parameters checked against the
engine). We report A and B as run and use the difference as a natural experiment (Section~\ref{sec:consistency});
the controlled experiment below adds a vLLM sampler with every decoder linear on the grid.

\paragraph{Statistics.} Each arm was run once. Where we quote a Welch $t$ or a standard error, it is computed over
the steps of one run and measures how well two arms separate relative to step-to-step fluctuation; consecutive steps
are strongly autocorrelated (lag-1 autocorrelation of \mt{rollout\_corr/kl} is 0.68--0.99 in setting B), so these are
not confidence levels from independent repeats. Because the arms of a setting see the same prompts at each step, we
also compare them step by step; at step~1 they also hold the same weights. ``Last 10'' means steps 51--60 (A), 23--32
(B), 21--30 (C) and 1--10 (D). \mt{chi2\_token} is spike-dominated in some runs and is summarized by its median.

\paragraph{Controlled experiments.} Two further groups of runs test the main claims with everything else held fixed.
They use setting A's configuration, software and host type (Appendix~\ref{app:controlled}). The \emph{matching
experiment} runs four pairings in vLLM at learning rate 0, so every arm keeps the checkpoint's weights at every step
and sees the same 64 prompts per step. The pairings are \pair{BF16}{BF16}, \pair{\mxf}{BF16}, \pair{\mxf}{\mxf} with
every decoder linear on the grid, and \pair{\mxf}{\mxf} with the exclusion bug F2 reproduced on purpose, which leaves
\mt{gate\_up\_proj} in BF16. Its statistics pair the arms by prompt and resample prompts (steps 2--10, 576 prompts), so its
intervals describe prompt-to-prompt variation within one run per arm. The \emph{micro-batch sweep} first records the
trajectories of eight steps and then replays them, so every run trains on identical data. It varies the update
stage's micro-batch over 1, 2, 4 and 8 sequences (old log-prob: twice as many) and adds a half batch, for a BF16 and
an \mxf{} learner at learning rate $10^{-6}$; two more runs profile the 1-sequence size with Nsight Systems.

\section{Results: consistency}
\label{sec:consistency}

\begin{figure}[t]
\centering
\includegraphics[width=\linewidth]{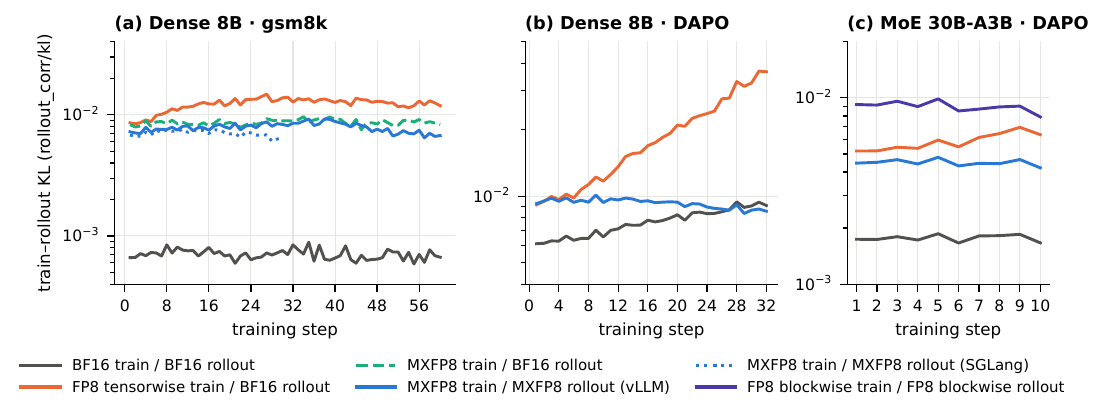}
\caption{Train--rollout KL per step (log scale). In (a) and (b) the vLLM \pair{\mxf}{\mxf} sampler kept
\mt{gate\_up\_proj} in BF16 (F2); the SGLang sampler in (a) and the MoE sampler in (c) put every decoder linear on the
learner's grid.}
\label{fig:kl}
\end{figure}

\begin{figure}[t]
\centering
\includegraphics[width=\linewidth]{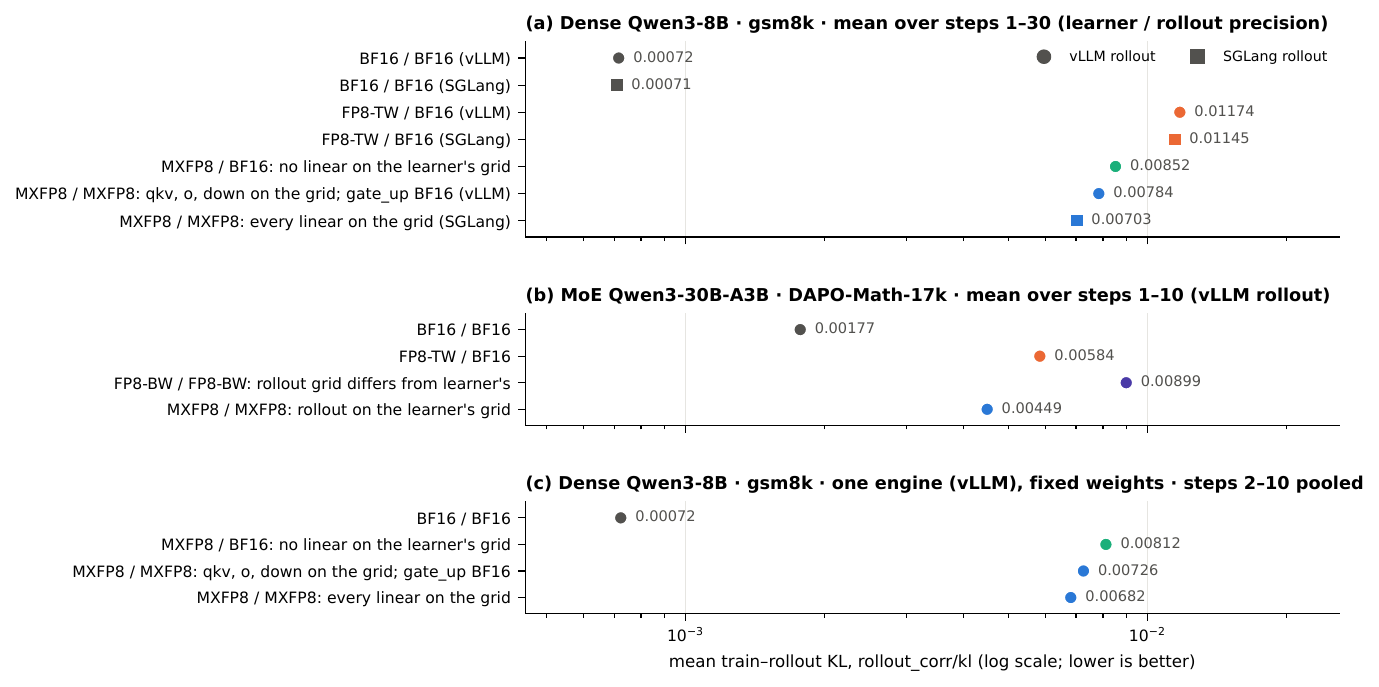}
\caption{Mean train--rollout KL by learner/sampler pairing. (a) For the same \mxf{} learner recipe, KL is lower when
more of the sampler's linear layers are on the learner's grid (none $\rightarrow$ qkv/o/down $\rightarrow$ all); the fully
matched sampler is SGLang, the other two are vLLM. The two control pairings, whose samplers are BF16, agree across the
two engines within 2.5\%. (b) On the MoE model, the shared-grid pipeline has half the KL of the FP8-blockwise
pipeline, whose sampler quantizes with FP32 scales while its learner is restricted to power-of-two scales. (c) The
controlled experiment in one engine, with the checkpoint's weights at every step and KL pooled over steps 2--10, gives
the same order as (a); Table~\ref{tab:matching} gives the paired reductions with their intervals.}
\label{fig:matching}
\end{figure}

Figure~\ref{fig:kl} shows the train--rollout KL of every arm, Figure~\ref{fig:matching} summarizes it by pairing, and
Table~\ref{tab:consistency} gives the four diagnostics. No divergence in these runs comes from stale data: trajectory
staleness is 0 at every step of every arm, so all of it is numerical.

\paragraph{Step 1 is a fixed-weight comparison.} The arms of a setting start from the same checkpoint with the same
seed and data order, and log their first diagnostics before any update. At step~1 they therefore hold the same
weights, see the same prompts, and differ only in precision. Their sampled responses differ, but each arm's
\mt{kl} estimates the KL under its own sampler, so this does not bias the comparison. After step~1 the weights
diverge, and a difference between two arms mixes their numerics with their training trajectories.
Table~\ref{tab:consistency} therefore reports step~1 next to the late-training means.

\begin{table}[t]
\centering
\small
\caption{Train--rollout mismatch. ``Step 1'' is \mt{kl} at the first step, where all arms of a setting hold the same
weights and see the same prompts; the other columns are last-10-step means (whole-run means in parentheses); lower
is better except for the Pearson correlation. $\chi^2$ is the median. Setting C's two control arms are omitted; their
step-1 KL is 0.00064 (\pair{BF16}{BF16}) and 0.00881 (\pair{\tw}{BF16}), and over the common steps 1--30 their KL
agrees with A's within 2.5\%.}
\label{tab:consistency}
\resizebox{\linewidth}{!}{%
\begin{tabular}{@{}lllllll@{}}
\toprule
Setting & Pairing & \mt{kl}, step 1 & \mt{kl} & \mt{log\_ppl\_abs\_diff} & Pearson & \mt{chi2\_token} \\
\midrule
\multirow{4}{*}{A: dense, gsm8k (vLLM)}
 & \pair{BF16}{BF16} & 0.00066 & 0.00069 (0.00071) & 0.00143 (0.00129) & 0.9954 (0.9965) & 0.0015 (0.0014) \\
 & \pair{\tw}{BF16} & 0.00859 & 0.01217 (0.01225) & 0.01317 (0.01325) & 0.9380 (0.9464) & 0.0404 (0.0408) \\
 & \pair{\mxf}{BF16} & 0.00828 & 0.00864 (0.00865) & 0.00953 (0.00940) & 0.9678 (0.9724) & 0.0220 (0.0230) \\
 & \pair{\mxf}{\mxf}, gate\_up BF16 & 0.00723 & 0.00693 (0.00784) & 0.00784 (0.00862) & 0.9791 (0.9758) & 0.0170 (0.0200) \\
\midrule
C: dense, gsm8k (SGLang) & \pair{\mxf}{\mxf}, all matched & 0.00681 & 0.00670 (0.00703) & 0.00730 (0.00759) & 0.9806 (0.9792) & 0.0158 (0.0183) \\
\midrule
\multirow{3}{*}{B: dense, DAPO (vLLM)}
 & \pair{BF16}{BF16} & 0.00610 & 0.00882 (0.00764) & 0.00864 (0.00749) & 0.9828 (0.9870) & 0.0224 (0.0184) \\
 & \pair{\tw}{BF16} & 0.00915 & 0.02969 (0.01907) & 0.02938 (0.01886) & 0.8833 (0.9424) & 0.2538 (0.0822) \\
 & \pair{\mxf}{\mxf}, gate\_up BF16 & 0.00927 & 0.00878 (0.00929) & 0.00862 (0.00914) & 0.9898 (0.9886) & 0.0198 (0.0211) \\
\midrule
\multirow{4}{*}{D: MoE, DAPO (vLLM)}
 & \pair{BF16}{BF16} & 0.00174 & 0.00177 & 0.00179 & 0.9964 & 0.0036 \\
 & \pair{\tw}{BF16} & 0.00519 & 0.00584 & 0.00595 & 0.9877 & 0.0125 \\
 & \pair{\bw}{\bw}, different grids & 0.00921 & 0.00899 & 0.00910 & 0.9854 & 0.0225 \\
 & \pair{\mxf}{\mxf}, all matched & 0.00446 & 0.00449 & 0.00457 & 0.9926 & 0.0103 \\
\bottomrule
\end{tabular}}
\end{table}

\begin{table}[t]
\centering
\small
\caption{Matching in one engine at fixed weights: vLLM, learning rate 0, setting A's configuration. ``Step 1'' is
\mt{kl} at the first step; the other columns pool steps 2--10 (576 prompts $\times$ 8 samples per arm, the same prompts
in every arm), so \mt{chi2\_token} and the Pearson correlation are taken over all tokens rather than as medians of
steps. The KL reductions carry 95\% intervals from 10\,000 bootstrap resamples of prompts, paired across arms.}
\label{tab:matching}
\resizebox{\linewidth}{!}{%
\begin{tabular}{@{}llllllll@{}}
\toprule
& & & & & & \multicolumn{2}{c}{KL reduction (95\% interval) against} \\
\cmidrule(l){7-8}
Pairing & \mt{kl}, step 1 & \mt{kl} & \mt{log\_ppl\_abs\_diff} & Pearson & \mt{chi2\_token} & BF16 sampler & gate\_up BF16 \\
\midrule
\pair{BF16}{BF16} & 0.00067 & 0.00072 & 0.00108 & 0.9979 & 0.00129 & -- & -- \\
\pair{\mxf}{BF16} & 0.00854 & 0.00812 & 0.00868 & 0.9778 & 0.02279 & -- & -- \\
\pair{\mxf}{\mxf}, gate\_up BF16 & 0.00719 & 0.00726 & 0.00787 & 0.9794 & 0.02013 & 10.6\% (8.6--12.7) & -- \\
\pair{\mxf}{\mxf}, all matched & 0.00697 & 0.00682 & 0.00739 & 0.9809 & 0.01841 & 16.1\% (14.1--18.1) & 6.1\% (3.9--8.3) \\
\bottomrule
\end{tabular}}
\end{table}

\paragraph{Finding 1: among low-precision pipelines, the shared-grid pipeline is the most consistent in every
setting; most of its late-training margin over per-tensor FP8 is that recipe's drift.} Against \pair{\tw}{BF16}, the
\pair{\mxf}{\mxf} pipeline lowers last-10-step KL by 43\% (A), 70\% (B), 46\% (C) and 23\% (D); the other three
diagnostics agree in direction in every setting. At step~1 the margins are smaller: 16\% (A), 23\% (C) and 14\% (D),
and none in B (0.00927 vs.\ 0.00915). Most of the late margin therefore comes from the growth of the per-tensor
pipeline's mismatch during training (Finding~4). BF16 end to end remains the most consistent pipeline by a wide margin
on the dense model (about 10$\times$ lower KL than any quantized pipeline in A) and by 2.5$\times$ on the MoE model.
The fp8 KV cache used by all arms of B raises the BF16 floor there (0.00610 at step~1), and over the last 10 steps
BF16 reaches the level of the \mxf{} pipeline.

\paragraph{Finding 2: for an MXFP8 learner, a sampler on the learner's grid is more consistent than a BF16 sampler,
and more so the more of its linear layers are on the grid.} The controlled experiment changes only the sampler: one
engine, identical weights at every step (learning rate 0), the same prompts and the same learner recipe
(Table~\ref{tab:matching}). Putting a layer on the grid changes its weights, the quantization of its input
activations and its GEMM kernel at once, so the experiment measures the recipe as a whole and does not isolate the
weight grid's share. Over steps 2--10, KL falls from 0.00812 with a BF16 sampler to 0.00726 when
\mt{gate\_up\_proj} stays off the grid (the configuration of F2) and to 0.00682 with every decoder linear on the grid,
10.6\% and 16.1\% below the BF16 sampler; the fully matched sampler is 6.1\% below the partially matched one (95\%
interval 3.9--8.3\%). The k3 estimator of the same divergence gives 10.2\%, 5.9\% and 15.5\%, with every interval
above zero. The log-perplexity gap and the Pearson correlation agree in direction and size. The token-level
$\chi^2$ agrees in direction, but its intervals include zero for each of the two single steps of matching (none to
half, half to all). The step-1 values are close to those of setting A's original runs (0.00854, 0.00719 and 0.00067
against 0.00828, 0.00723 and 0.00066). Over training, setting A shows the same order for the partially matched
sampler: it is below the BF16 sampler at each of the first 10 steps (by 0.00102, or 14\%, on average) and at 29 of the
first 30, and over the last 10 steps the gap is 20\% (0.00693 vs.\ 0.00864; Welch $t=11.4$). In setting C, SGLang's
fully matched sampler has a step-1 KL of 0.00681, 18\% below A's BF16 sampler, and the ordering none $>$ partial $>$
full holds at 29 of the first 30 steps (Figure~\ref{fig:matching}a). That cross-engine reduction is close to the
same-engine 16\%, consistent with a small engine effect, although the two come from different runs and hosts. Because
the synchronized weights equal the learner's byte for byte in every layer we compared, and the two quantizers agree on
every weight matrix (Section~\ref{sec:bytes}), the mismatch that remains with every layer matched most likely comes
from the computation, the activations and the kernels, rather than from the stored weights
(Section~\ref{sec:discussion}).

\paragraph{Finding 3: an FP8 pipeline whose sampler is on a different grid is less consistent than the shared-grid
pipeline, from the first step.} In setting D the FP8-blockwise pipeline quantizes both sides, but its learner uses
power-of-two block scales (required on Blackwell, where TE emulates the recipe with MXFP8 GEMMs) while its sampler
uses FP32 scales and $1\times128$ FP32-scaled activations. Its KL is twice the shared-grid pipeline's both at step~1
(0.00921 vs.\ 0.00446), before the runs diverge, and over the last 10 steps (0.00899 vs.\ 0.00449). It is also 54\%
above FP8 tensorwise training with a BF16 sampler (0.00584), a comparison that changes the learner recipe as well.
The blockwise and shared-grid pipelines differ in three ways: the learner recipe, the sampler's weight grid, and the
sampler's activation quantization. These runs do not separate the three. Per layer, the blockwise and MXFP8 learners
are equally accurate on B200: with power-of-two scales, \mt{te.Linear} and \mt{te.GroupedLinear} give the same errors
against BF16 under both recipes to four digits (Appendix~\ref{app:blockwise}). That points away from the learner's
own accuracy, but it is not decisive. Equal error magnitudes do not make equal errors, and the mismatch depends on how
the learner's error relates to the sampler's (Section~\ref{sec:discussion}), which a per-layer accuracy test does not
measure. A related mismatch at 688B scale, an MXFP8 learner with a block-wise FP8 sampler, has been reported to
collapse \citep{skt2026axk2}.

\paragraph{Finding 4: per-tensor FP8 drifts; block-scaled FP8 does not.} The \pair{\tw}{BF16} mismatch grows during
training: in B its KL rises 3.1$\times$ from the first to the last eight steps (0.0100 $\rightarrow$ 0.0313), in A by
27\%, and in D by 25\% within ten steps. The \mxf{} pipeline stays flat (B: 0.0095 $\rightarrow$ 0.0087; A: $-7\%$; D:
$-2\%$). A per-tensor scale is set by the largest element of the whole tensor; as RL sharpens the policy and
activation outliers grow, more of each tensor is represented coarsely, and the learner drifts from its BF16
sampler. A 32-element block scale is local. This explanation fits all three settings, but we did not measure the
activation distributions that would confirm it.

\paragraph{The recipe/matching split depends on the window.} Along the path
\pair{\tw}{BF16}\,$\rightarrow$\allowbreak\,\pair{\mxf}{BF16}\,$\rightarrow$\allowbreak\,\pair{\mxf}{\mxf} in setting A, with the partially matched
vLLM sampler, the share of the KL reduction that falls on changing the learner recipe is 23\% at step~1, 53\% over
steps 1--10, 67\% over the last 10 steps and 82\% over the whole run; the rest falls on matching the sampler. At
identical weights matching is the larger part, and later the per-tensor learner's drift (Finding~4) dominates. These
shares are arithmetic along one path through three runs, not a causal decomposition. On tail diagnostics the recipe
dominates: the matched and unmatched \mxf{} arms have similar dual-clip trigger rates, both about a third of \tw's. A
plausible reading is that matching cancels a systematic offset between the two sides, while the per-token spread that
drives the importance-weight variance is set mostly by the learner's scaling granularity and by activation
quantization, which matching does not touch.

\section{Results: speed and memory}
\label{sec:speed}

\subsection{Why the obvious timing metrics mislead}
\begin{figure}[t]
\centering
\includegraphics[width=\linewidth]{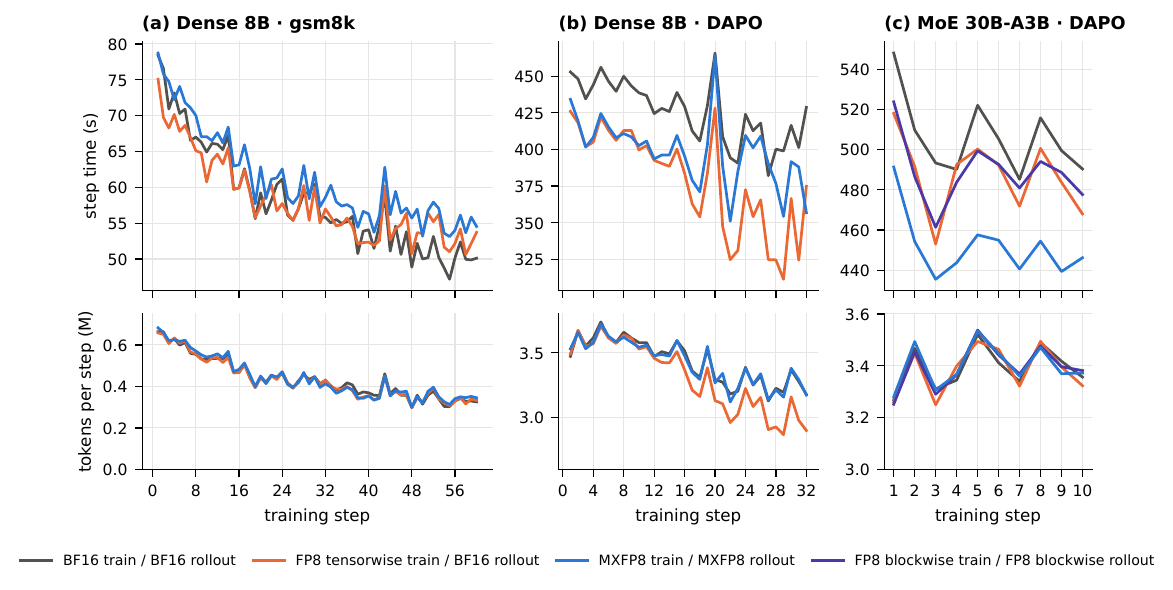}
\caption{Step time (top) and tokens per step (bottom) of every arm; each column ran on one host. (a) Dense 8B,
gsm8k: step time follows the tokens per step, which halve as responses shorten; MXFP8 is slower than BF16 at 57 of
60 steps. (b) Dense 8B, DAPO: MXFP8 is faster than BF16 at 31 of 32 steps at equal token counts; the per-tensor arm
is faster still but generates fewer tokens. (c) MoE 30B-A3B: token counts agree with BF16's within 0.2\%, and MXFP8
is the fastest arm at every step. Per-stage curves are in Appendix~\ref{app:stagetime}.}
\label{fig:steptime}
\end{figure}

In RL the number of tokens per step changes as the policy learns. In setting A the MXFP8 arm's mean response
shortens from 1\,247 tokens at step 1 to 599 over the last ten steps, the tokens per step $T$ of every arm roughly
halve, and step time follows them (Figure~\ref{fig:steptime}). Step time is therefore a function of the learning
trajectory, and per-token time $t/T=F/T+c$ is contaminated by any fixed cost $F$, which grows in relative weight as $T$
shrinks. In setting A per-token \mt{update\_actor} time ranks BF16 fastest, while the per-token cost $c$ fitted below
ranks it slowest by 74\% (Appendix~\ref{app:method}). We therefore fit each stage's time as $t=F+c\,T$ across the 57
post-warm-up steps of one run, using the variation of $T$ that RL provides for free. The reference-model pass is BF16
in every arm and serves as a control: its fitted $c$ agrees across the three arms within 1.4\% (10.38, 10.40,
10.26\ms/token).

\subsection{The added fixed cost is a cost per micro-batch}
\label{sec:permb}
Table~\ref{tab:fits} gives the fits. Both FP8 recipes cut the per-token cost of the learner's GEMM-heavy stages
(\mt{update\_actor} by 45\% and 43\%) but add a token-independent cost (+5.8\,s and +6.6\,s per step). Within one
stage the number of micro-batches is the same at every step, so a single fit cannot tell a cost paid once per step
from one paid per micro-batch. Three observations favor the second. First, the old-log-prob pass runs 64
forward-only micro-batches per step and \mt{update\_actor} runs 128 forward--backward micro-batches over the same 512
sequences, and the ratio of their added fixed costs is 0.117 for both recipes. A per-micro-batch cost reproduces this
ratio if a forward--backward micro-batch carries about 4.3 times the overhead of a forward-only one, which is
plausible because the backward pass quantizes output gradients for two GEMMs per layer and the training forward also
produces the transposed copies that wgrad needs; this is a consistency check, not an independent measurement. For
\mxf{} the implied overhead is 51\,ms per forward--backward and 12\,ms per forward-only micro-batch, on the order of
0.1\,ms per quantized GEMM call. SkyRL made the same observation on H100, where repeating the quantization work across
64 micro-batches put its trainer on a host-dispatch-limited path~\citep{skyrl2026fp8}. Second, the added cost depends
on the host: on the slower-CPU host of setting C, \mt{update\_actor} time barely depends on $T$ at all
($R^2=0.02$--0.31), the stage is bound by host-side work, and both FP8 recipes are \emph{slower} than BF16 (+23\% and
+31\%). Third, the per-micro-batch form predicts a held-out workload with a different micro-batch count better than a
per-step form does (Section~\ref{sec:breakeven}).

\paragraph{A controlled test.} The micro-batch sweep removes the confound: it replays the same token data at every
step and changes only the micro-batch size, of the update and of the old-log-prob pass alike (Table~\ref{tab:sweep});
the learners still train at $10^{-6}$, which changes the weights' values but not the work. The two forms then predict different things. A per-step cost
leaves the difference between \mxf{} and BF16 the same at every micro-batch size of the full batch, about $-0.9$\,s
per step with setting A's constants. A per-micro-batch cost makes the difference grow with the number of micro-batches, from
$-4.2$\,s at 64 to $+18.9$\,s at 512. The measured difference follows the second form: $-7.61$, $+0.64$, $+5.81$ and
$+15.37$\,s per step at 64, 128, 256 and 512 micro-batches. Fitted on the sweep alone,
$t=a+N_{\text{mb}}\,\delta+c\,T$ gives an added cost of $\Delta\delta=47.0\pm4.5$\,ms per micro-batch (standard
errors), against 51.5\,ms inferred from setting A. Its added cost per step, $\Delta a=-0.10\pm3.21$\,s (95\% interval
$-6.4$ to $+6.2$\,s), is too loose to rule out a per-step part by itself; the evidence for the per-micro-batch form is
how the difference grows with the number of micro-batches. Padding is a small part of
it: \mxf{} pads each sequence to a multiple of 32 tokens and each micro-batch to a multiple of 256, which adds 10.6\%
tokens at one sequence per micro-batch, and fitting on padded tokens lowers $\Delta\delta$ only to 44.0\,ms. The
weights are quantized once per mini-batch rather than per micro-batch (below), so their quantization is a per-step
cost, too small to register in $\Delta a$. The old-log-prob stage gives $\Delta\delta=24.2\pm6.3$\,ms per forward-only
micro-batch, twice the 12.0\,ms inferred from setting A.

\begin{table}[t]
\centering
\small
\caption{Micro-batch sweep on identical replayed data: update-stage time per step (mean of steps 2--6) and the
difference between the learners, against the predictions of a per-micro-batch and a per-step cost with setting A's
constants ($\Delta\delta=51.5$\,ms, $\Delta c=11.36\,\mu$s/token, $\Delta F=6.58$\,s); the two predictions coincide
at 128 micro-batches, setting A's own count, from which $\Delta\delta$ was derived. $\pm$ is the standard error over
the five paired steps. Bottom: fits of $t=a+N_{\text{mb}}\,\delta+c\,T$ to the 25 points of each learner, with standard
errors.}
\label{tab:sweep}
\resizebox{\linewidth}{!}{%
\begin{tabular}{@{}lrrrrrr@{}}
\toprule
\multicolumn{2}{@{}l}{Micro-batch} & & & & \multicolumn{2}{c}{Predicted \mxf{} $-$ BF16 (s)} \\
\cmidrule(r){1-2}\cmidrule(l){6-7}
sequences (tokens) & $N_{\text{mb}}$ per step & BF16 (s) & \mxf{} (s) & \mxf{} $-$ BF16 (s) & per micro-batch & per step \\
\midrule
1 (1.3k) & 512 & 69.12 & 84.49 & $+15.37\pm0.44$ & $+18.9$ & $-0.9$ \\
2 (2.6k) & 256 & 42.90 & 48.71 & $+5.81\pm0.97$ & $+5.7$ & $-0.9$ \\
4 (5.2k) & 128 & 32.72 & 33.36 & $+0.64\pm0.91$ & $-0.9$ & $-0.9$ \\
8 (10.3k) & 64 & 32.47 & 24.86 & $-7.61\pm1.68$ & $-4.2$ & $-0.9$ \\
4, half batch (5.2k) & 64 & 17.43 & 16.55 & $-0.88\pm0.75$ & $-0.5$ & $+2.8$ \\
\midrule
\multicolumn{7}{@{}l}{Fit, BF16: $a=0.32\pm2.80$\,s, $\delta=85.9\pm3.9$\,ms, $c=35.6\pm5.0\,\mu$s/token} \\
\multicolumn{7}{@{}l}{Fit, \mxf: $a=0.22\pm1.58$\,s, $\delta=132.9\pm2.2$\,ms, $c=24.0\pm2.8\,\mu$s/token} \\
\multicolumn{7}{@{}l}{Difference: $\Delta a=-0.10\pm3.21$\,s, $\Delta\delta=47.0\pm4.5$\,ms,
$\Delta c=-11.6\pm5.8\,\mu$s/token, $m^\ast=4.0$k tokens (4.1k with setting A's $\Delta c$)} \\
\bottomrule
\end{tabular}}
\end{table}

\paragraph{Where the time goes.} Nsight Systems traces of the 1-sequence runs split the added cost
(Table~\ref{tab:profile}). Per micro-batch of the update stage, \mxf{} launches 912 more kernels: 325 quantization
kernels for activations and gradients, and 576 kernels that rearrange block scales into the layout the GEMM reads.
TE already writes that layout directly for most activations and gradients (180 of the 325 quantization kernels do).
The rearrangements serve the tensors it keeps in the compact layout: the cached weights, whose scales are rearranged
again for the forward and backward GEMMs of every micro-batch (288 kernels), and the four tensors per layer that
sequence parallelism all-gathers across the tensor-parallel pair (the inputs of the qkv and first MLP projections and
the output gradients of the other two), each read by two GEMMs (288), because the all-gather concatenates compact
scales. These add 4.8\,ms of GPU time, and the NCCL kernels run
5.1\,ms longer (\mxf{} all-gathers the FP8 data and its scales as separate operations, and an NCCL kernel's time
includes waiting for the other rank), but the FP8 GEMMs save 7.1\,ms, so summed kernel time grows by only 3.5\,ms of
the 43.4\,ms added. Summed kernel time counts kernels that overlap on different streams twice, and BF16 overlaps more
of them (3.4\,ms against 0.9\,ms); with kernels and memory copies merged across streams, the GPU is busy for 5.9\,ms
more, 14\% of the added time. For the other 37.4\,ms no GPU operation runs at all. Matching each such gap to the CUDA
call that issued the operation ending it shows why: for 30.6\,ms the host had not yet issued the next operation,
6.3\,ms fall between an operation's issue and its start, and only 0.5\,ms pass with the next operation issued but
waiting on other work. While the GPU waited for the host, the issuing thread was outside every traced call for nearly all of
the added 30.6\,ms: its time in TE's C++ calls and in NCCL calls grows by 1.6\,ms each, its time in CUDA API calls
falls by 2.7\,ms (the BF16 run spends 4.5\,ms per micro-batch mapping memory), and synchronizing calls take 1.0\,ms in
both recipes. The added cost is therefore host-side work between GPU operations, in code the trace does not
instrument (Python, PyTorch's dispatcher and autograd, and the Python layers of TE and Megatron), rather than GPU work
or waiting for the GPU; the forward-only old-log-prob pass shows the same pattern. At this micro-batch size the GPU is
busy for only 44\% of the stage under BF16 and 37\% under \mxf. The profiler slows the host (the stage takes 81.5\,s
and 103.7\,s under it, against 69.1\,s and 84.5\,s without) but not the kernels; without it the difference is
30.0\,ms per micro-batch, of which the added GPU time would be about a fifth. Only rank 0 was traced, and which
functions carry the host time remains to be measured (Section~\ref{sec:future}). This is the kind of cost SkyRL
observed on H100~\citep{skyrl2026fp8}. The trace also shows that
the weights are quantized once per call rather than once per micro-batch: besides its per-micro-batch instances, a cast
kernel runs 144 more times per call, once for each linear layer on a rank (288 times per step in the update stage,
which makes two calls, and 144 in the forward-only old-log-prob pass). Two remedies follow, neither tested here.
Writing the cached weights' scales in the GEMM's layout when they are quantized would remove the 288 weight
rearrangements per micro-batch, about 1.8\,ms of profiled host time at 6.2\,$\mu$s each; TE~2.18, newer than the
2.16.1 we used, does this by default for \mxf{} weights~\citep{te2026preswizzle}. The other 288 remain as long as the
all-gather needs compact scales. Capturing the micro-batch in a CUDA graph would remove most of the host dispatch.

\begin{table}[t]
\centering
\small
\caption{Where the added cost goes: update stage with 1-sequence micro-batches, per micro-batch (Nsight Systems,
step 3, rank 0, 512 micro-batches). Kernel times are summed over all streams. GPU busy time merges kernels and memory
copies across streams; the rest of the wall time, with no GPU operation running, is split by when the host issued the
operation that ends each gap. The profiler inflates host-side time. The last row is from the timed runs without the
profiler.}
\label{tab:profile}
\begin{tabular}{@{}lrrr@{}}
\toprule
Per micro-batch & BF16 & \mxf & Difference \\
\midrule
Wall time under the profiler (ms) & 159.3 & 202.6 & $+43.4$ \\
\quad GPU kernel time, summed (ms) & 72.0 & 75.5 & $+3.5$ \\
\qquad GEMM & 20.4 & 13.3 & $-7.1$ \\
\qquad \mxf{} quantization & -- & 2.2 & $+2.2$ \\
\qquad scale rearrangement & -- & 2.6 & $+2.6$ \\
\qquad NCCL & 13.3 & 18.3 & $+5.1$ \\
\qquad attention, normalization, other & 38.3 & 39.0 & $+0.7$ \\
\quad GPU busy, merged across streams (ms) & 69.6 & 75.5 & $+5.9$ \\
\quad no GPU operation running (ms) & 89.7 & 127.1 & $+37.4$ \\
\qquad next operation not yet issued by the host & 68.8 & 99.4 & $+30.6$ \\
\qquad issued, not yet started (launch latency) & 16.4 & 22.7 & $+6.3$ \\
\qquad issued earlier, waiting on other work & 4.6 & 5.1 & $+0.5$ \\
Kernel launches & 4\,139 & 5\,051 & $+912$ \\
Wall time without the profiler (ms) & 135.0 & 165.0 & $+30.0$ \\
\bottomrule
\end{tabular}
\end{table}

\begin{table}[t]
\centering
\small
\caption{Stage cost model $t=F+c\,T$ in setting A (fast host, steps 4--60), and the break-even point per micro-batch.
$\Delta\delta=\Delta F/N_{\text{mb}}$ with $N_{\text{mb}}=128$ (update) and 64 (old log-prob); $m^\ast=\Delta\delta/\Delta c$.
$\pm$ are standard errors: of the fit for $F$ and $c$, and for $m^\ast$ from the delta method with the covariance of
each fit's intercept and slope (correlation $-0.98$). A bootstrap over steps gives 95\% intervals for $m^\ast$ of
4.3--4.8k (\mxf) and 3.6--3.9k (\tw) for \mt{update\_actor}, and 2.1--5.6k and 1.6--5.1k for the old log-prob, whose
break-even point lies at the edge of the measured range.}
\label{tab:fits}
\resizebox{\linewidth}{!}{%
\begin{tabular}{@{}lllllll@{}}
\toprule
Stage & Arm & $F$ (s) & $c$ ($\mu$s/token) & $R^2$ & $\Delta\delta$ (ms per micro-batch) & $m^\ast$ (tokens per micro-batch) \\
\midrule
\multirow{3}{*}{\mt{update\_actor} (fwd+bwd)} & BF16 & $12.31\pm0.30$ & $26.74\pm0.69$ & 0.964 & -- & -- \\
 & \tw & $18.10\pm0.28$ & $14.60\pm0.66$ & 0.899 & 45.3 & $3\,726\pm61$ \\
 & \mxf & $18.89\pm0.24$ & $15.38\pm0.56$ & 0.932 & 51.5 & $4\,527\pm108$ \\
\midrule
\multirow{3}{*}{\mt{old\_log\_prob} (fwd)} & BF16 & $3.40\pm0.24$ & $10.60\pm0.55$ & 0.872 & -- & -- \\
 & \tw & $4.08\pm0.12$ & $8.06\pm0.27$ & 0.942 & 10.6 & $4\,161\pm666$ \\
 & \mxf & $4.17\pm0.23$ & $8.03\pm0.53$ & 0.809 & 12.0 & $4\,668\pm701$ \\
\midrule
\multirow{3}{*}{\mt{ref} (BF16 in all arms; control)} & BF16 & $2.70\pm0.09$ & $10.38\pm0.22$ & 0.977 & -- & -- \\
 & \tw & $2.74\pm0.09$ & $10.40\pm0.21$ & 0.978 & -- & -- \\
 & \mxf & $2.79\pm0.09$ & $10.26\pm0.20$ & 0.980 & -- & -- \\
\bottomrule
\end{tabular}}
\end{table}

\subsection{A break-even point in tokens per micro-batch}
\label{sec:breakeven}
\begin{figure}[t]
\centering
\includegraphics[width=\linewidth]{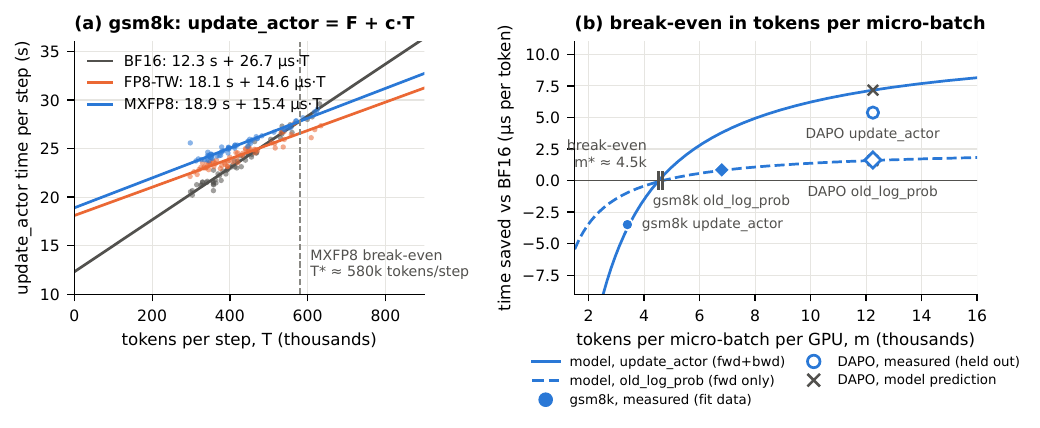}
\caption{(a) \mt{update\_actor} time per step against tokens per step in setting A, with least-squares fits. (b) Time
saved per token by \mxf{} against tokens per micro-batch: Eq.~\eqref{eq:breakeven} with constants from setting A
(curves), measurements from A (filled) and from the held-out setting B (open), and the model's prediction for B
($\times$).}
\label{fig:breakeven}
\end{figure}

If a recipe adds $\Delta\delta$ per micro-batch and saves $\Delta c$ per token, a micro-batch of $m$ tokens saves
$\Delta c\,m-\Delta\delta$, and the recipe pays off when
\begin{equation}
  m \;>\; m^\ast = \frac{\Delta\delta}{\Delta c}.
  \label{eq:breakeven}
\end{equation}
For \mxf{} on this model and host, $m^\ast$ is 4.5k tokens for the forward--backward stage (95\% bootstrap interval
4.3--4.8k). The forward-only estimate is similar (4.7k) but loosely determined (2.1--5.6k), because with 64
micro-batches its break-even point falls at the smallest step size we measured. A per-step threshold such as ``580k
tokens per step'' is only $m^\ast$ multiplied by a particular micro-batch count. Equation~\eqref{eq:breakeven} is
supported four ways (Figure~\ref{fig:breakeven} and Table~\ref{tab:sweep}).
\begin{itemize}
\item \textbf{Under control.} In the sweep the same data and learner run at 1.3k, 2.6k, 5.2k and 10.3k tokens per
micro-batch. \mxf{} is 15.37 and 5.81\,s per step slower at the two smallest sizes, within 0.64\,s of BF16 at 5.2k, and
7.61\,s faster at 10.3k, so the break-even point lies close to 5k, near the 4.5k fitted on setting A. Fitted on the
sweep alone, $m^\ast$ is 4.0k tokens, or 4.1k with setting A's per-token saving, and resampling the five steps puts
it between 2.8k and 7.6k. Two caveats apply. The sweep varies
$T$ only through its half-batch run, so it determines $\Delta c$ poorly: without the 8-sequence size the fit gives
$\Delta c=-2.6\pm4.3\,\mu$s/token and $\Delta\delta=38.2\pm3.6$\,ms. And BF16 does not get faster from 4 to 8
sequences per micro-batch; its 8-sequence point lies 6.8\,s above the line fitted without it, while \mxf{} stays on
its line, so part of \mxf's lead at 10.3k is BF16's slowdown. BF16's 8-sequence steps also vary most (standard
deviation 4.0\,s, against 0.6--2.5\,s in the other runs), and between adjacent sizes, where the tokens per step are
equal, each extra micro-batch adds 37.3 and 40.4\,ms up to 5.2k tokens but 128.9\,ms from 5.2k to 10.3k.
\item \textbf{Within one run.} In setting A, \mt{update\_actor} micro-batches carry about 3.4k tokens, below
$m^\ast$, and the stage is 6.9\% \emph{slower} under \mxf{}; old-log-prob micro-batches carry about 6.8k tokens, above
$m^\ast$, and that stage is 4.0\% \emph{faster}, in the same run. The update stage also crosses $m^\ast$ as
responses shorten (Appendix~\ref{app:stagetime}). In the first four steps, at 4.9--5.3k tokens per micro-batch,
\mxf{} is 2.9\% faster than BF16; over the last ten, at 2.7k, it is 13.6\% slower. These are the steps the fit
uses, so the crossing illustrates Eq.~\eqref{eq:breakeven} rather than testing it.
\item \textbf{On held-out data.} Setting B uses dynamic batching with a budget of 12\,288 tokens per micro-batch.
Each batch is split into $\lceil\text{tokens}/12\,288\rceil$ micro-batches, so they carry 12.2k tokens on average;
this follows from the batching rule and the logged token counts, since per-micro-batch sizes were not logged. With
the constants fitted on A, Eq.~\eqref{eq:breakeven} predicts per-token savings of
9.0\% (old log-prob) and 15.9\% (\mt{update\_actor}) for \mxf; the measured savings are 9.1\% and 11.9\%. For \tw{} the
predictions are 9.4\% and 18.7\% against 10.5\% and 15.0\%. A model with a per-step fixed cost predicts 13.2\% and
20.9\% for \mxf, missing by 4.1 and 9.0 points.
\item \textbf{Across hosts.} On the slower host the same micro-batches do not break even and FP8 training is a net
loss (above), as expected if the fixed cost is host-side. There, time barely depends on $T$, so $m^\ast$ cannot be
fitted.
\end{itemize}
The remaining 4-point over-prediction for \mt{update\_actor} on B may reflect effects that the two-parameter model
ignores, such as a smaller FP8 advantage for the backward GEMMs at long sequence lengths; we did not profile that
workload. The model also suggests how $m^\ast$ scales, which we have not tested: the per-token saving grows with the
per-GPU GEMM size (hidden dimension, and inversely with tensor parallelism), while the per-call cost does not, so wider
models or lower tensor parallelism should break even at smaller micro-batches. Because the per-micro-batch cost is mostly
host-side (Section~\ref{sec:permb}), CUDA graphs or a faster host should lower $m^\ast$
directly; writing the
cached weights' scales in the GEMM's layout, as TE~2.18 does, removes only a small part of it.

\subsection{End-to-end step time}
\begin{figure}[t]
\centering
\includegraphics[width=\linewidth]{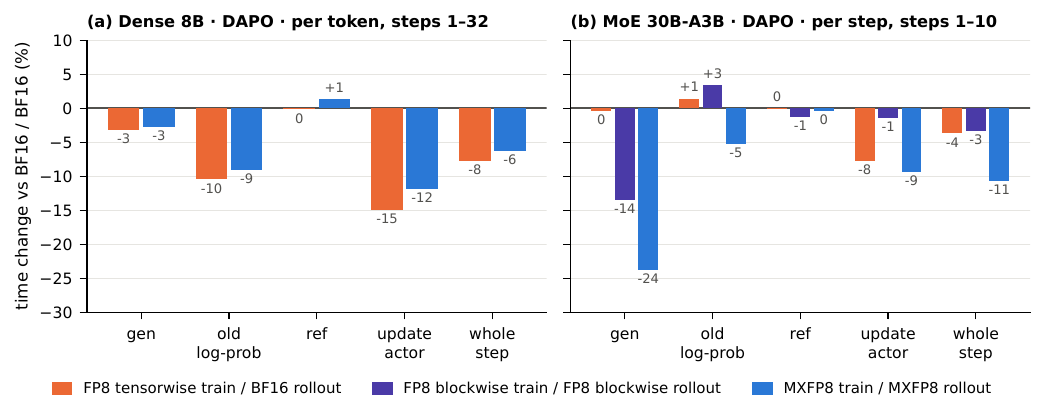}
\caption{Stage time relative to \pair{BF16}{BF16} on the same host. (a) Dense, setting B, per token, because the
\tw{} arm generated 3.5\% fewer tokens. (b) MoE, setting D, per step (tokens per step within 0.2\% across arms).}
\label{fig:stages}
\end{figure}

Figure~\ref{fig:stages} shows per-stage changes against BF16 in the two long-sequence settings; Table~\ref{tab:moe}
gives setting D in full.

\begin{table}[t]
\centering
\small
\caption{Setting D (Qwen3-30B-A3B, 4 B200, one host): mean over 10 steps. Actor memory is the allocator
high-water mark.}
\label{tab:moe}
\begin{tabular}{@{}lrrrr@{}}
\toprule
& \pair{BF16}{BF16} & \pair{\tw}{BF16} & \pair{\bw}{\bw} & \pair{\mxf}{\mxf} \\
\midrule
step (s) & 506.0 & 487.2 ($-3.7\%$) & 488.9 ($-3.4\%$) & \textbf{451.8 ($-10.7\%$)} \\
\quad generation & 133.6 & 133.1 ($-0.4\%$) & 115.5 ($-13.5\%$) & \textbf{101.7 ($-23.8\%$)} \\
\quad old log-prob & 63.9 & 64.8 ($+1.4\%$) & 66.1 ($+3.4\%$) & 60.5 ($-5.3\%$) \\
\quad reference (BF16) & 57.3 & 57.3 ($0.0\%$) & 56.5 ($-1.3\%$) & 57.1 ($-0.4\%$) \\
\quad update\_actor & 244.4 & 225.4 ($-7.8\%$) & 240.8 ($-1.5\%$) & 221.3 ($-9.4\%$) \\
\quad weight sync & 3.5 & 3.4 & 6.3 & 7.5 \\
throughput (tokens/s/GPU) & 1\,678 & 1\,738 & 1\,738 & 1\,883 ($+12.2\%$) \\
actor peak memory (GiB) & 60.0 & 67.2 & 74.6 & 75.0 \\
mean response length & 6\,476 & 6\,461 & 6\,483 & 6\,491 \\
\bottomrule
\end{tabular}
\end{table}

\paragraph{Dense.} In setting B the \mxf{} pipeline lowers step time by 6.4\% (\mt{update\_actor} $-12.1\%$, old
log-prob $-9.3\%$) and is faster than BF16 at 31 of 32 steps (Figure~\ref{fig:steptime}b). The per-tensor arm is
faster per step still ($-11.0\%$), partly because it generates fewer tokens: 3.5\% fewer on average and 9\% fewer by
step~32. Per token it is 7.8\% faster than BF16, against 6.3\% for \mxf, because its quantization is cheaper
(Table~\ref{tab:fits}). Its sampler is BF16, and its mismatch grows during training (Finding~4). In setting A, where
\mt{update\_actor} micro-batches are below $m^\ast$, \mxf{} raises step time by 5.1\% and is slower than BF16 at 57 of
60 steps. Generation is essentially unchanged in both (per token $-1.1\%$ in A, $-2.7\%$ in B). Because these vLLM
samplers kept \mt{gate\_up\_proj} in BF16, they do not measure the dense MXFP8 rollout speed; the SGLang runs do not
either, because SGLang~0.5.12 dispatches MXFP8 dense GEMMs to a generic Triton kernel by default, which made
generation 1.9$\times$ slower than BF16 (52 vs.\ 27\,ms per response token; its FlashInfer CUTLASS backend,
usable with weight synchronization after our loader fix, reduces the gap to 1.3$\times$, 34\,ms per token). The controlled matching runs
do measure it, with every decoder linear on the grid and FlashInfer autotuning on: over steps 2--10 the fully matched
sampler generates 5.3\% more tokens per second than the BF16 sampler of the same learner and 2.0\% more than the BF16
pipeline. The two BF16 samplers, identical except for the learner they serve, differ from each other by 3.1\%, so at
this model size and batch dense MXFP8 rollout shows no clear speed-up.

\paragraph{MoE.} In setting D the \mxf{} pipeline is the fastest arm in every stage that runs the model in low
precision: step time $-10.7\%$, generation $-23.8\%$, \mt{update\_actor} $-9.4\%$, at identical response lengths,
and the fastest arm at each of the ten steps (Figure~\ref{fig:steptime}c).
These times include parameter offloading and untuned FlashInfer kernels in every arm, so they are comparisons
between arms, not the best achievable times. The
FP8-blockwise pipeline speeds up generation by 13.5\% but its training barely at all ($-1.5\%$), consistent with TE
emulating block scaling through MXFP8 GEMMs plus a conversion of every block-scaled operand on this hardware. We did
not trace why MoE generation benefits and dense generation (in these configurations) did not; MoE decode is
dominated by expert GEMMs whose weight traffic FP8 halves, which would explain it.

\paragraph{Weight synchronization.} Quantizing the weights at every synchronization costs about 1\,s per step on the
dense model (3.2 $\rightarrow$ 4.2\,s in A) and 4\,s on the MoE model including the self-check (3.5 $\rightarrow$
7.5\,s), 1--2\% of a step. The cost is per step and does not shrink with batch size. Timers inside the vLLM worker
show where it goes in the controlled dense runs, where a synchronization takes 3.1\,s with \mxf{} against 1.4\,s
without: TE's quantization of all 252 linear weights uses about 11\,ms of GPU time, and most of the added time is
spent restoring the quantized parameters to a loadable layout before the transfer (0.57\,s) and re-deriving the
kernels' layouts after it (0.49\,s).

\subsection{Memory}
FP8 training costs memory: TE keeps FP8 copies and scales next to the BF16 weights. The actor's peak allocation grows
by 4.7\% in A and 2.2\% in B, where activations dominate, and by 25\% in D (60.0 $\rightarrow$ 75.0\,GiB); the
FP8-blockwise recipe costs the same in D (74.6\,GiB), so the cost belongs to TE's FP8 training rather than to the MXFP8
format.

\section{Results: learning dynamics}
\label{sec:learning}

\paragraph{Reward and accuracy.} Over these short horizons no pipeline shows a quality regression that the data can
resolve (Table~\ref{tab:quality}). In setting A, held-out gsm8k accuracy after 60 steps is 0.9393 (\pair{\mxf}{\mxf}),
0.9363 (BF16), 0.9356 (\pair{\mxf}{BF16}) and 0.9249 (\tw); repeating one configuration on another host moved accuracy
by up to 2.0 points in an earlier 20-step round, and the \mxf{} sampler scores 1.9 points higher than the BF16
samplers on the \emph{untrained} model, so these differences are not an ordering. In setting B, reward, response
length and truncation rate of \pair{\mxf}{\mxf} track BF16 in every 8-step block (last 10 steps: score 0.135 vs.\ 0.150,
Welch $t=0.4$), while \pair{\tw}{BF16} reaches a higher training reward (0.240) together with 7\% shorter responses,
five times BF16's KL to the reference policy and the growing train--rollout mismatch above; without a held-out set we
cannot tell faster learning from early over-exploitation. Ten steps of setting D resolve no quality difference.

\begin{table}[t]
\centering
\small
\caption{Learning signals. gsm8k accuracy is on the full test set (1\,319 problems) at step 0 and at the last step.
Entropy is the mean of steps 1--3 against the last 10 steps (D: steps 8--10).}
\label{tab:quality}
\resizebox{\linewidth}{!}{%
\begin{tabular}{@{}llllll@{}}
\toprule
Setting & & \pair{BF16}{BF16} & \pair{\tw}{BF16} & \pair{\mxf}{BF16} & \pair{\mxf}{\mxf} \\
\midrule
A & gsm8k accuracy, step 0 $\rightarrow$ 60 & 0.3169 $\rightarrow$ 0.9363 & 0.3169 $\rightarrow$ 0.9249 & 0.3154 $\rightarrow$ 0.9356 & 0.3359 $\rightarrow$ 0.9393 \\
A & entropy change & $-49\%$ & $-47\%$ & $-23\%$ & $-4\%$ \\
C & gsm8k accuracy, step 0 $\rightarrow$ 30 & 0.3184 $\rightarrow$ 0.8870 & 0.3230 $\rightarrow$ 0.8946 & -- & 0.3738 $\rightarrow$ 0.8795 \\
C & entropy change & $-34\%$ & $-32\%$ & -- & $+4\%$ \\
B & score, last 10 steps & 0.150 & 0.240 & -- & 0.135 \\
B & entropy change & $-31\%$ & $-74\%$ & -- & $+4\%$ \\
D & entropy change (\pair{\bw}{\bw}: $+9\%$) & $-7\%$ & $-19\%$ & -- & $+6\%$ \\
\bottomrule
\end{tabular}}
\end{table}

\begin{figure}[t]
\centering
\includegraphics[width=\linewidth]{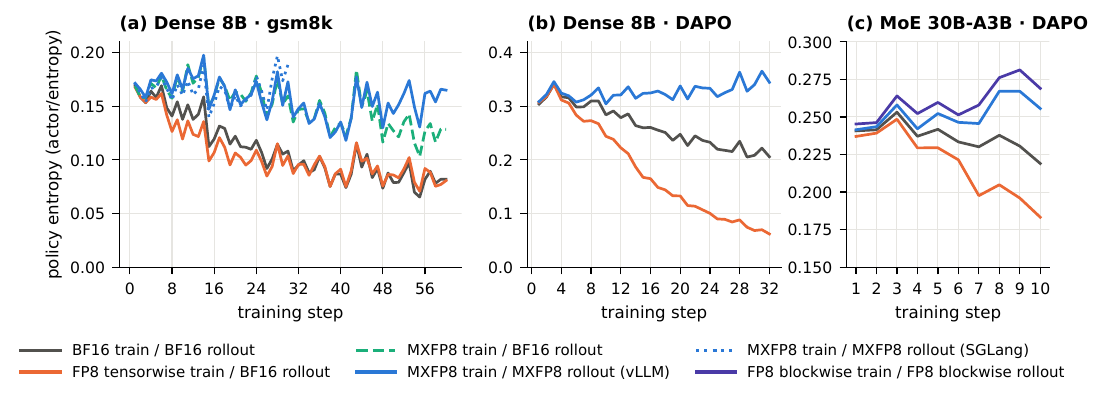}
\caption{Policy entropy per step. Under \mxf{} training the entropy stays flat or rises; under BF16 and \tw{} it
decays, fastest under \tw{} on DAPO-Math-17k ($-74\%$ in 32 steps).}
\label{fig:entropy}
\end{figure}

\paragraph{Entropy.} The learning signal that separates the recipes most clearly is the policy's entropy
(Figure~\ref{fig:entropy}). Under BF16 and \tw{} it decays, as is usual for RL on reasoning tasks; under \mxf{} it
stays flat or rises slightly, in both dense tasks, on both engines, and on the MoE model. On DAPO-Math-17k the \tw{}
policy's entropy falls by 74\% in 32 steps, while its training reward is the highest of the three arms. Three
observations constrain the explanation. The effect is
not a measurement artifact: at steps 1--3 the \mxf{} learner reads the same entropy as BF16, and in the ablation arm
the BF16 sampler, evaluating the \mxf-trained weights on its own samples, sees the same slower decline (its
log-perplexity falls by 30\%, against 54\% for the BF16 pipeline), so the flatter policy is encoded in the weights. It is specific to the
recipe: \tw{} decays like BF16 on gsm8k and faster on DAPO. And both sides contribute: the ablation arm, with an
\mxf{} learner and a BF16 sampler, lands in between ($-23\%$). QeRL reports that quantization noise raises policy entropy and aids exploration~\citep{huang2025qerl}, and
full-pipeline FP8 RL has been observed to produce entropy surges~\citep{chen2026fullfp8}; the sign of the effect
therefore depends on the recipe, and we see the direction here but not the mechanism. E4M3
rounding is multiplicative, about 3.6\% RMS per element, and the same magnitude of error under the \tw{} recipe moves
entropy the other way, so it is not simply noise acting as temperature. What the flatter entropy means for training
is open. Reward and held-out accuracy ended close to BF16's (Table~\ref{tab:quality}), so the \mxf{} policy did
learn. Whether the retained entropy is useful exploration or slower sharpening, and whether the faster \tw{}
decline is quicker learning, over-exploitation, or the onset of the entropy collapse known from RL on reasoning
tasks~\citep{yu2025dapo,cui2025entropy}, needs longer runs with held-out evaluation. It does mean that step-matched
checkpoints from different recipes are not at the same point of training.

\section{Correctness at the train--inference boundary}
\label{sec:failures}

Bringing the pipeline up on B200 surfaced seven failures (Table~\ref{tab:failures}). They share a pattern: the
synchronization reports success and every shape and dtype matches, yet the sampler no longer computes the learner's
policy. In five of the seven (F1--F4, F7) the job also exits with code~0 and logs normal-looking metrics. Five (F1,
F3--F6) appear only on the target hardware, because on pre-Blackwell GPUs vLLM serves MXFP8 through Marlin or
emulation fallbacks and the native kernels never run. Rather than add one check per failure, we state what a
synchronization must preserve, test those conditions against the running engine, state what each test can and
cannot see, and exercise the tests with injected violations.

\subsection{What a weight synchronization must preserve}
After every synchronization the sampler must satisfy seven conditions:
\begin{itemize}[leftmargin=2.6em]
\item[C1] \emph{Layer set.} It quantizes exactly the layers whose GEMMs the learner runs in low precision, and keeps
the others in high precision.
\item[C2] \emph{Weight grid.} Its quantized weights are the learner's forward weight grid.
\item[C3] \emph{Derived state.} Every kernel-specific copy of weights or scales derived at load (swizzled scales,
shuffled or interleaved experts, packed layouts) is rebuilt from the synchronized weights.
\item[C4] \emph{Storage identity.} The tensors that CUDA graphs captured are the ones that receive the update.
\item[C5] \emph{Completeness.} When the last bucket of a streamed update has been loaded, every parameter has been
written and every staged buffer has been folded back.
\item[C6] \emph{Output sanity.} Its token probabilities are finite, and its divergence from the learner is bounded.
\item[C7] \emph{Observability.} Every check reports a verdict, or reports why it could not run.
\end{itemize}
Each failure in Table~\ref{tab:failures} breaks one of these. In every silent one the synchronization returned
success and every tensor kept its shape and dtype, so no check of those could have failed.

\begin{table}[t]
\centering
\small
\caption{Failures found on B200 and the condition each breaks. ``Silent'' means the job exits 0 with metrics that are
logged normally.}
\label{tab:failures}
\resizebox{\linewidth}{!}{%
\begin{tabular}{@{}>{\raggedright\arraybackslash}p{0.5cm}>{\raggedright\arraybackslash}p{3.1cm}>{\raggedright\arraybackslash}p{1.1cm}>{\raggedright\arraybackslash}p{4.4cm}>{\raggedright\arraybackslash}p{4.3cm}>{\raggedright\arraybackslash}p{3.3cm}@{}}
\toprule
& Failure & Breaks & Symptom & Root cause & Caught now by; fix \\
\midrule
F1 & \mt{lm\_head} quantized in the sampler (silent) & C1 & reward 0.000 for all 20 steps; responses at the length cap; NaN probability difference; untrained model scores 0.000 (healthy: 0.316) & not isolated: excluding \mt{lm\_head} removes the NaN logits; value range, scale handling in the vocabulary-parallel head and the kernel path remain candidates; with NaN logits EOS is never sampled & audit, sentinel; \mt{lm\_head} and embedding kept in BF16 \\
F2 & router exclusion silently removes dense MLP layers (silent) & C1 & none in the metrics; 56 of 311 parameters flagged on Qwen3-1.7B & vLLM~0.24's ModelOpt matcher falls back to substring matching, so \mt{mlp.gate} also excludes \mt{mlp.gate\_up\_proj} & audit (engine re-check); exact matching on module paths \\
F3 & stale swizzled scales after sync, SGLang FlashInfer CUTLASS (silent) & C3 & held-out accuracy 0.000 from step 0; KL 7.2; entropy 10.4 ($\approx$ uniform over the vocabulary) & the backend derives an interleaved scale copy at load that the sync never rebuilds & self-check, sentinel; refit loader re-runs post-processing \\
F4 & stale MoE expert layout after sync, vLLM TRT-LLM MXFP8 MoE (silent) & C3 & expert probe relative error 1.74 at the first sync (healthy 0.05); KL 0.047 vs.\ 0.0022 on a small randomly initialized MoE & an in-place repack that keeps shape and dtype escaped shape-based detection; a ``done'' flag also blocked re-processing & self-check (MoE probe); record same-shape repacks, merged as verl~\#7986 \\
F5 & BF16 MoE sync crashes on SM100 & C3 & crash at the first sync & vLLM picks TRT-LLM's unquantized MoE kernel, whose 4-D layout the BF16 per-expert sync cannot write & crash; in-place staging, merged as verl~\#7987 \\
F6 & FP32 block scales on SM100 & C2 & assertion at the first FP8 quantization & TE emulates block scaling with MXFP8 GEMMs, whose E8M0 scales keep only the exponent & assertion; FP32 scales documented as Hopper-only, verl~\#8062 \\
F7 & monitors that do not report (silent) & C7 & audit ran but printed nothing & a transitive dependency raised the root logger to ERROR in worker processes; the log level was not forwarded to Ray actors & dedicated handler and forwarded level; trace mode; warning when a check cannot run \\
\bottomrule
\end{tabular}}
\end{table}

\subsection{The checks and what they can see}
Table~\ref{tab:checks} maps the guards of Section~\ref{sec:system}, and the offline byte comparison of
Section~\ref{sec:bytes}, to the conditions. Two design choices matter more
than any single check. First, each check reads the engine's own state rather than re-deriving the answer from the
configuration or from the synchronization's success. The audit asks the engine which parameters it holds in FP8, and
the self-check runs the serving kernel. The rule-based comparison missed F2, and every check that does not run the
kernel missed F4. Second, divergence alarms cannot replace these mechanism checks. F2's KL (0.007--0.010) lies inside
the range of healthy quantized pipelines, so no threshold separates it. F4's broken expert layout gave 0.047, less than 30\%
above the highest per-step KL of any arm we ran (0.037, per-tensor FP8 in setting B). The sentinel therefore
keeps a threshold of 1 for gross failures and leaves moderate faults to the mechanism checks.

\begin{table}[t]
\centering
\small
\caption{Checks, the conditions they test, and their coverage by construction. All but the byte comparison run
during training.}
\label{tab:checks}
\resizebox{\linewidth}{!}{%
\begin{tabular}{@{}>{\raggedright\arraybackslash}p{2.4cm}>{\raggedright\arraybackslash}p{1.1cm}>{\raggedright\arraybackslash}p{2.8cm}>{\raggedright\arraybackslash}p{5.0cm}>{\raggedright\arraybackslash}p{5.4cm}@{}}
\toprule
Check & Tests & Runs & Catches & Cannot see \\
\midrule
quantized-layer audit & C1 & once, at the first sync after a training step & any parameter whose precision differs between learner and engine, in either direction, read from the engine's live state (vLLM) & values (C2--C6); the engine query counts only linear and expert layers, so a quantized \mt{lm\_head} shows only in the rule half; where no engine query exists (SGLang, FP8-blockwise path) it compares with the configured rule and inherits the rule's errors \\
self-check & C3 (C6 for the probed layers) & every sync & stale derived state in every layer of a kind, whatever its shape and dtype, because it reads the kernel's output & C2 (kernel and reference use the same synchronized weights); faults confined to layers or experts it does not probe; C4 (it calls the kernel eagerly, not through the CUDA graph) \\
step sentinel & C6 & every step & non-finite probabilities, KL above 1, nearly every response truncated on two steps, whatever the cause & divergence below its thresholds (F2, F4) \\
refit-cycle assertions & C4, C5 & every sync & an update that cannot be folded into the captured storage; staged buffers not folded back; scales a streamed update never wrote (SGLang) & storage the cycle does not track \\
byte comparison (offline) & C2, C3 & at two synchronizations of the controlled runs & a sampler weight, scale or kernel layout that differs from TE's quantization of the synchronized weights, or a learner forward weight that does (Section~\ref{sec:bytes}) & layers it does not sample (3 of 36); it is not run during training \\
\bottomrule
\end{tabular}}
\end{table}

\subsection{The weight grid, checked byte for byte}
\label{sec:bytes}
C2 holds by construction when both sides run the same quantizer on the same weights, but the two quantizations are
separate calls in separate processes on differently laid out tensors. The learner's forward quantizes Megatron's fused,
tensor-parallel-sharded parameters (qkv interleaved by key-value group, gate and up stacked); the synchronization
quantizes the exported tensors inside the vLLM process; and the engine then derives its kernel layouts. A step that
rounded, cast or misplaced values would produce a slightly different grid without an error, and the self-check cannot
see it, because its reference uses the same synchronized weights. We therefore compared the bytes in two of the
controlled runs: at the 11th synchronization of the fully matched run of the matching experiment (learning rate 0), and at the 7th
synchronization of an \mxf{} sweep run, after seven training steps (14 optimizer updates) at $10^{-6}$ had changed
5.77\% of the compared BF16 elements. Those changes alter 0.84\% of the FP8 weight bytes (4.9 million of 579 million,
at least 0.6\% in each of the 24 matrices) and no scale, so a sampler still holding the step-0 weights would have
differed from TE's quantization in every compared matrix (Appendix~\ref{app:controlled}). For layers 0, 17 and 35, all four projections and both
tensor-parallel ranks, we captured three things: the engine's E4M3 weights and E8M0 scales after the last bucket and
before post-processing; the layouts the kernel reads after post-processing (a transposed view and swizzled scales
under FlashInfer's CuTe-DSL kernel); and the learner's cached forward weights at the start of the next update, which
its old-log-prob pass had just used. The engine's bytes equal TE's quantization of the synchronized weights, the
kernel layouts hold the same bytes, and the learner's cached weights equal the same quantization: in both runs, 0 of
about 600\,MB differ in each comparison. TE's quantizer and vLLM's own MXFP8 quantizer also agree byte for byte on all
252 linear weight matrices of Qwen3-8B (6.95 billion elements) and on boundary cases such as block maxima of exactly
$448\times2^k$. For this format, therefore, a shared grid is the same as a shared format with a conforming quantizer.
Because vLLM's MXFP8 linear kernel quantizes its input activations with that same function (only the scale layout
differs), the two sides also quantize activations by the same rule (Section~\ref{sec:discussion}).

\subsection{Evidence from real faults}
Table~\ref{tab:detection} shows which check catches each of the four silent faults that reached a training run. The
entries are measured where the check or its signal was present in the run, reproduced by a CPU test, or follow from
the check's construction. The checks are complementary. Without the audit, F2 would have gone undetected, and
without the self-check, F4 would have. The sentinel overlaps with the other two on F1 and F3, but it is the only
check that needs no model of the failure mechanism. The audit's rule-based half alone would also have missed F2:
only asking the engine exposed it.

\begin{table}[t]
\centering
\small
\caption{Which check catches each silent fault that reached a run. m: measured on B200; t: reproduced by a CPU test;
c: follows from the check's construction.}
\label{tab:detection}
\resizebox{\linewidth}{!}{%
\begin{tabular}{@{}>{\raggedright\arraybackslash}p{3.4cm}>{\raggedright\arraybackslash}p{3.0cm}>{\raggedright\arraybackslash}p{3.0cm}>{\raggedright\arraybackslash}p{3.3cm}>{\raggedright\arraybackslash}p{3.3cm}@{}}
\toprule
Fault & audit, rule half & audit, engine re-check & self-check & sentinel \\
\midrule
F1 \mt{lm\_head} quantized & caught (t) & missed: the engine query counts only linear and expert layers (c) & missed: \mt{lm\_head} is not probed (c) & fires: NaN probability difference (m) \\
F2 \mt{gate\_up\_proj} left in BF16 & missed: 28 of 28 layers agree (m) & caught: 56 of 311 parameters (m) & missed: probed layers are healthy (c) & silent: KL 0.007--0.010, far below 1 (m) \\
F3 stale swizzled scales (SGLang) & missed: layer set correct (c) & no engine query (c) & caught (t) & fires: KL 7.2 (m) \\
F4 stale expert layout (vLLM) & missed: 2 of 2 layers agree (m) & missed: 69 of 69 parameters agree (m) & caught: 1.739 vs.\ 0.25 at the first sync (m) & silent: KL 0.047 for 4 steps (m) \\
\bottomrule
\end{tabular}}
\end{table}

Two deliberate trips on hardware confirm that each stop path works end to end. With its tolerance lowered to
$10^{-6}$, the MoE self-check raised at the first synchronization. With the truncation trigger set to one step, the
sentinel stopped a randomly initialized MoE whose responses all hit the length cap at step~1. In the healthy MXFP8
runs (setting D and the small-MoE validation runs), the self-check read 0.026--0.027 on the dense probe and
0.049--0.052 on the expert probe. That is five to nine times below its tolerance of 0.25, against 1.739 for F4. The
audit agreed with the engine on all 69 and all 18\,867 parameters it checked, and neither check raised a false alarm
in these runs. We observed one false alarm: on the FP8-blockwise path, where the audit has no engine query, its rule
assumes that \mt{lm\_head} is quantized, so it reported one disagreement in setting D. vLLM's FP8 method never
quantizes the output projection. On the MoE model, a synchronization with quantization and the per-synchronization checks takes
4\,s longer than a BF16 one, about 1\% of a step. We did not separate the checks' share from the quantization's. On the
dense model, quantization alone adds 1\,s; the dense runs predate the checks.

\subsection{Reproducible fault tests and an injection protocol}
The checks and the refit cycle ship with 108 CPU tests in verl, in nine files under \mt{tests/\allowbreak utils}.
They run without a GPU against stub engines that reproduce the vLLM and SGLang kernel layouts. They inject
violations, for example:
\begin{itemize}
\item a scale layout left stale after the weights change;
\item a repack that keeps shape and dtype;
\item an engine that keeps a layer in BF16 against the configured rule;
\item a quantized output projection;
\item scales that a streamed update never wrote;
\item a staged buffer that was never folded back;
\item non-finite probabilities.
\end{itemize}
These tests check that each violation is caught. Other tests in the same files check that healthy configurations
raise nothing. F1--F5 are reproduced this way as violations of their conditions. F6 is a loud configuration
assertion, and F7's logging interaction is fixed in code but not under test. The conditions and the injections are
not specific to verl: any framework that updates a quantized sampler in place must preserve C1--C7.
Table~\ref{tab:injection} is the hardware protocol that extends the CPU tests. For each injection and check, it
reports:
\begin{itemize}
\item detection and miss rates;
\item false alarms on clean runs;
\item the number of synchronizations from injection to detection;
\item the added synchronization time and peak memory;
\item each check's own contribution, measured by repeating the injections with that check disabled.
\end{itemize}
So far we have the CPU column and the hardware evidence above. The full hardware campaign has not been run
(Section~\ref{sec:limitations}).

\begin{table}[t]
\centering
\small
\caption{Fault-injection protocol: injected violation, the condition it breaks, the check expected to catch it, and
the evidence available now.}
\label{tab:injection}
\resizebox{\linewidth}{!}{%
\begin{tabular}{@{}>{\raggedright\arraybackslash}p{5.0cm}>{\raggedright\arraybackslash}p{1.1cm}>{\raggedright\arraybackslash}p{4.6cm}>{\raggedright\arraybackslash}p{5.2cm}@{}}
\toprule
Injected violation & Breaks & Expected to catch it & Evidence now \\
\midrule
a class of layers left in BF16 by the sampler & C1 & audit, engine re-check & CPU tests; hardware (F2) \\
new weights with stale derived scales & C3 & self-check & CPU tests; hardware (F4) \\
content permuted, shape and dtype kept & C3 & self-check, which reads the output & CPU tests; hardware (F4) \\
corruption confined to one expert or a late layer & C3 & none, by construction & open: size of the blind spot \\
update written outside the CUDA-graph storage & C4 & the refit cycle's fold; the eager self-check cannot see it & CPU tests of the fold; open: graph replay \\
update incomplete across buckets & C5 & refit-cycle assertions & CPU tests (SGLang loader, BF16 MoE staging); open on hardware \\
non-finite or diverging sampler output & C6 & sentinel & CPU tests; hardware trip \\
a check that runs but reports nothing & C7 & warning when a check cannot run; trace mode & CPU tests in part; hardware (F7) \\
\bottomrule
\end{tabular}}
\end{table}

\section{Discussion}
\label{sec:discussion}

\paragraph{Why matching helps.} The learner and the sampler compute the same network with different numerics. For a
low-precision learner, one can split their difference into a weight-grid term, an activation-quantization term, and a
kernel term (accumulation order, fused operations, attention) that is present even in BF16. Writing $e_t$ and $e_r$
for the learner's and the sampler's output errors against a common high-precision reference, the mismatch is
$\|e_t-e_r\|^2=\|e_t\|^2+\|e_r\|^2-2\langle e_t,e_r\rangle$, so it depends on how the two errors correlate as well as on
their sizes. A BF16 sampler of an MXFP8 learner has $e_r\approx0$ and inherits the learner's entire weight and
activation quantization error as mismatch. A sampler on the learner's grid makes the weight parts of the two errors
identical, so they cancel, and leaves the activation term, which the two engines compute with different kernels on
similar inputs. Two quantizers with equal error sizes but independent rounding cancel nothing, which is why equal
per-layer accuracy does not make two pipelines equally consistent (Finding~3). Our measurements fit this picture:
matching lowers the systematic offset (KL, log-perplexity) more than the per-token spread (importance-weight variance,
dual-clip rate), and in one engine at identical weights KL falls step by step as more of the sampler's linear layers
join the learner's grid, by 16\% in all. Each step moves a layer's weights onto the learner's grid and its activations
onto the learner's quantization rule at once, so the experiment does not split the reduction between the two terms. With the weights byte-identical where we checked them, about 84\% of the BF16-sampler
mismatch remains, and
it does not come from different quantization rules: vLLM quantizes activations with its own MXFP8 function, which
gives the same bytes as TE's quantizer on every weight and on boundary cases (Section~\ref{sec:bytes}). It most likely comes from that rule acting on slightly different inputs, which the two engines' BF16
operations (attention, normalization, accumulation order) produce, and from those operations themselves. SkyRL
reports the same structure: with bitwise-identical weights a per-token log-probability gap of 0.02--0.03 remains, which
it attributes to activation quantization and kernel differences~\citep{skyrl2026fp8}, and QUADS finds activation error
to be the dominant remaining source for NVFP4~\citep{zhuge2026quads}. Sharing the activations' quantized values
themselves, which requires bitwise-identical upstream kernels, would be the next step; the BF16 floor shows how much
remains.

\paragraph{Why MXFP8 on Blackwell.} On B200, block scaling is native only in MX form. The FP8-blockwise recipe is
emulated through MXFP8 GEMMs, so it gains little training speed (Table~\ref{tab:moe}), and because its scales must
then be powers of two, the FP32-scaled weights of verl's existing FP8 rollout no longer match its learner's grid.
MXFP8 removes both problems: the learner's recipe is the hardware's native one, and the sampler can receive the
identical grid. On Hopper, where cuBLASLt applies FP32 block scales directly, unified block-wise pipelines
\citep{jetrl2026,miles2025unifiedfp8,nemorl2026fp8} are the corresponding matched choice; SkyRL's ablation of FP32
against power-of-two block scales there found a slightly larger gap for power-of-two scales and no difference in
convergence~\citep{skyrl2026fp8}.

\paragraph{Guidance for practitioners.}
\begin{enumerate}
\item Quantize the sampler with the learner's quantizer, and keep \mt{lm\_head}, embeddings and routers in high
precision. Check the layer set (C1) against the engine's live parameters, not against the configured rule.
\item After every synchronization, run the serving kernel of at least one layer of each kind against a reference built
from the synchronized weights (C3). A healthy layer's error is a few percent; a stale layout's is $O(1)$. Do not rely
on divergence alarms for this: a broken expert layout raised KL to only 0.047.
\item Estimate $m^\ast$ from one short run by regressing stage time on tokens per step, and, if possible, repeat with a
second micro-batch size to separate per-step from per-micro-batch cost; then choose micro-batch token budgets well
above $m^\ast$. The per-micro-batch cost is mostly host-side work between GPU operations, so a faster host or CUDA
graphs should lower it; TE~2.18 or later also stops rearranging the cached weights' scales in every micro-batch. Do not benchmark FP8 RL at small
micro-batches, and compare timings only on one host.
\item On Blackwell, prefer MXFP8 to the FP8-blockwise recipe, which needs power-of-two scales there.
\item Monitor \mt{actor/entropy} when changing recipes, and do not treat step-matched checkpoints from different
recipes as equivalent.
\end{enumerate}

\subsection{Limitations}
\label{sec:limitations}
\begin{itemize}
\item \textbf{One seed, short runs.} Each arm ran once: the four settings for 10--60 steps at learning rate $10^{-6}$,
the controlled runs for 3--12 steps (learning rate 0 in the matching experiment). Our within-run
statistics are not confidence intervals; the controlled experiments' intervals resample prompts within one run per
arm, not runs. We cannot rank final quality.
\item \textbf{Fixed-weight comparisons only at the initial checkpoint.} The controlled matching experiment compares
samplers at identical weights in one engine, but only at the checkpoint and each with its own samples; over training,
the fully matched dense data come from SGLang, which crosses engines. Evaluating trained checkpoints on fixed token
sequences through a matrix of learner and sampler precisions in one engine, with the exact KL over the vocabulary at
sampled positions, would extend it to trained weights and size the interaction with the fp8 KV cache of setting B.
Each sampler pairing switches weights, activation quantization and kernels together, so the experiment does not
separate the weight grid's share (Section~\ref{sec:future}).
\item \textbf{An FP8 KV cache in setting B.} To shorten generation of responses of up to 8192 tokens, every arm of B
ran vLLM with \texttt{kv\_cache\_dtype=fp8}. Keys and values are then stored as FP8 E4M3 with one scale per tensor,
1.0 because none is calibrated, and with vLLM~0.24's default attention backend on B200 the query is quantized to FP8
as well. This is not MXFP8 and not part of the pipelines under study; settings A, C and D and the controlled runs keep
the KV cache in BF16. The learner has no KV cache: it recomputes attention over the whole sequence from BF16 queries,
keys and values. Every sampler in B therefore carries a quantization source the learner lacks, whatever its weight
format. The arms of B share it, so the comparisons within B stand, but B's absolute KL is not comparable with that of
the other settings, and its BF16 arm is not BF16 end to end: its step-1 KL is 0.00610, against 0.00066 in A, a gap to
which B's data and response lengths also contribute. The cache's share was not measured; switching only
\texttt{kv\_cache\_dtype} at fixed weights, as in the controlled runs, would measure it.
\item \textbf{The byte comparison is sampled and offline.} It covers 3 of 36 layers of the dense model at two
synchronizations and is not run during training; the MoE path was not compared byte for byte.
\item \textbf{The break-even model rests on one model and one host type.} The sweep confirms the per-micro-batch form
on fixed data but determines the per-token saving poorly, because the tokens per step vary only through its half-batch
run, so its own $m^\ast$ spans 2.8--7.6k under resampling; BF16's largest micro-batch lies off the linear model; and
the forward-only stage's per-micro-batch cost is twice the value inferred from setting A. Each sweep configuration ran
once, BF16 before \mxf; the reference stage, the same work in every run, differs by 0.4--2.2\% within the full-batch
pairs and 8.2\% in the half-batch pair, so differences of a few percent are within run-to-run variation. The profile
covers one micro-batch size and one rank, the profiler inflates host time, and it does not attribute the added host
time to functions. The
standard errors of the setting-A fits assume independent residuals (their lag-1 autocorrelation is within $\pm0.2$).
The predicted scaling with model width and tensor parallelism is untested.
\item \textbf{The hardware fault-injection campaign has not been run.} The evidence for the checks is four real
faults, two deliberate trips, healthy runs, and CPU fault-injection tests. We have not measured on hardware the
detection and false-alarm rates under systematic injection, the detection delay, each check's time and memory cost, or
runs with each check disabled (Table~\ref{tab:injection}). The cases the checks cannot see by construction are
likewise not sized: faults confined to unprobed layers or experts, and storage the refit cycle does not track.
\item \textbf{MoE coverage.} One MoE family, vLLM only, ten steps; the FP8-blockwise path has no self-check. In
setting D the CPU optimizer's fused AdamW computes the weight-decay factor $1-\eta\lambda$ in FP32, where $1-10^{-8}$
rounds to 1, so weight decay was inactive in all arms (Appendix~\ref{app:incidental}).
\item \textbf{Rollout speed on dense models} is measured only in the short controlled runs, where the difference lies
within the spread between two identically configured BF16 samplers (F2 in vLLM and the Triton default in SGLang
prevented measuring it in settings A--C).
\item \textbf{What the entropy difference means} for final quality is unknown (Section~\ref{sec:learning}).
\end{itemize}

\subsection{Future experiments}
\label{sec:future}
Four experiments, each reusing the controlled runs' configuration and instrumentation, would close the main gaps above.
\begin{itemize}
\item \textbf{Splitting the matching gain.} In the matching experiment, three more samplers would compute every
decoder linear in BF16 but with the learner's MXFP8-quantized weights dequantized to BF16, with inputs rounded to MXFP8
and back, or with both. Compared with the BF16 and fully matched samplers, and with repeats of those two, they would
split the 16\% reduction between the weight grid, activation quantization and the FP8 kernels, and measure
run-to-run variation.
\item \textbf{Repeating the sweep near the break-even point.} Two more runs per learner at 4 and 8 sequences per
micro-batch, in alternating order and with allocator statistics, would show whether the crossing near 5k tokens and
BF16's slowdown at 8 sequences reproduce.
\item \textbf{Attributing the host time.} A CPU sampling profile with Python stacks on both ranks would name the
functions that carry the added host time. Repeating the 1-sequence runs without sequence parallelism, which removes
the FP8 all-gathers and the scale rearrangements they require, and with TE writing the cached weights' scales in the
GEMM's layout (the default from version 2.18), would measure two remedies.
\item \textbf{Comparing every layer byte for byte.} Extending the byte comparison from 3 to all 36 layers at one
synchronization would remove its sampling.
\end{itemize}

\section{Related work}
\label{sec:related}

\paragraph{Low-precision training.} FP8 formats and per-tensor scaling were introduced for deep learning
by~\citet{micikevicius2022fp8}; FP8-LM trained GPT-style models with FP8 GEMMs, gradients and optimizer
states~\citep{peng2023fp8lm}; DeepSeek-V3 used fine-grained block scaling and higher-precision accumulation at
671B scale~\citep{deepseekv3}. Microscaling formats were proposed by~\citet{rouhani2023mx} and standardized by the
OCP~\citep{ocp2023mx}; \citet{mishra2025mxfp8} give recipes for MXFP8 pre-training, and \citet{nvidia2025nvfp4}
pre-train with NVFP4. \citet{kumar2024precision} model how precision interacts with scale, and
\citet{khattak2025accum} model the accumulation behavior of NVIDIA tensor cores. These works concern one model
computed by one software stack; RL adds a second stack that must agree with the first.

\paragraph{Train--inference mismatch.} Differences between the sampler's and the learner's probabilities make RL
off-policy and have been corrected with truncated or masked importance sampling
\citep{yao2025offpolicy,liu2025speedkills}. \citet{qi2025fp16} reduce the mismatch by using FP16 on both sides;
MiniMax-M1 keeps the output head in FP32 for the same reason~\citep{minimax2025m1}; batch-invariant and deterministic
kernels remove the kernel-level part \citep{he2025nondeterminism,sglang2025deterministic}.

\paragraph{Low precision on one side.} FlashRL serves INT8 and FP8 rollouts of a BF16 learner and corrects the
resulting mismatch with truncated importance sampling~\citep{liu2025flashrl}; FP8-RL adds FP8 KV caches and a
production stack in verl~\citep{qiu2026fp8rl}. QeRL trains LoRA adapters over an NVFP4-quantized base that the sampler
shares, and reports that quantization noise raises entropy and aids exploration~\citep{huang2025qerl}.

\paragraph{Low precision on both sides.} Jet-RL keeps training and rollout on one block-wise FP8 precision flow on
H100 and reports that a BF16 learner with an FP8 sampler degrades on long rollouts~\citep{jetrl2026}; the LMSYS/Miles
unified FP8 pipeline~\citep{miles2025unifiedfp8} and NVIDIA's NeMo RL~\citep{nemorl2026fp8} do the same for MoE and
dense models, the latter reporting lower mismatch for end-to-end FP8 than for FP8 generation alone. On Blackwell, Miles
runs MXFP8 and NVFP4 end to end with a byte-exact quantizer contract checked by unit tests
\citep{miles2026mxfp8,miles2026report}, and SkyRL synchronizes the trainer's FP8 payloads directly so that the two
sides' weights are bitwise identical~\citep{skyrl2026fp8}. \citet{skt2026axk2} report reward collapse for an MXFP8
learner paired with a block-wise FP8 sampler. QUADS finds that for NVFP4 rollouts the weights can be aligned through a
shared quantization path while activation error dominates the remaining instability~\citep{zhuge2026quads}, and
\citet{chen2026fullfp8} address entropy surges in full-pipeline FP8 RL with calibrated clipping.

Our implementation is concurrent with the MXFP8 support in Miles and SkyRL and uses the same principle. What this
paper adds is measurement: how consistency depends on matching, at identical weights in one engine and over training,
including the case of a low-precision learner with a BF16 sampler, with the synchronized weights compared byte for
byte; when the speed-up materializes, as a threshold in tokens per micro-batch fitted on one workload, tested on a
held-out workload and in a controlled sweep, with a profile of the cost; and what a refit must preserve across the
train--inference boundary. For that last part we give run-time checks whose coverage we state, evidence of how they fared on real
faults, CPU fault-injection tests, and a protocol for measuring the checks on hardware.

\section{Conclusion}
\label{sec:conclusion}
RL can run in MXFP8 end to end on Blackwell, and several frameworks now do. Our measurements say three things about
that design. First, for a low-precision learner the sampler should share the learner's grid. In one engine at
identical weights, an MXFP8 sampler is more consistent than a higher-precision BF16 one, and more so the more of its
linear layers are on the learner's grid: 16\% lower KL with all of them, with the synchronized weights byte-identical to the
learner's in every layer we compared. An FP8 pipeline whose sampler uses a different grid has twice the mismatch of the shared-grid pipeline from
the first step. Without an FP8 KV cache, BF16 end to end remains the most consistent pipeline. Second, the speed-up is
conditional: the recipe adds a fixed cost per micro-batch, confirmed on fixed data, for most of which the GPU waits
for the host, so the gain appears only when each micro-batch carries more than a few thousand tokens and the
host is fast enough; long-sequence and MoE workloads are where it pays. Third, correctness across the train--inference
boundary can be stated as conditions and tested at run time. The failures we met each broke one condition, and most
were silent. Two faults occurred after the checks were in place. Both were caught by checks that read the engine's own
state, and both were missed by checks of the configuration, of shapes and of divergence. On the four silent faults
the checks complement one another rather than overlap. They cost about 1\% of a step on the MoE model, so they belong
in every synchronization, not only in a test suite.

\section*{Acknowledgements}
The implementation, the experiment design, the analysis and the drafting of this report were carried out with the
assistance of an AI coding assistant. Every line of code and text was reviewed by the
author, who takes full responsibility for the content.

\bibliographystyle{plainnat}
\bibliography{refs}

\appendix
\section{Measurement methodology}
\label{app:method}

\paragraph{Per-token time inverts the ordering.} In setting A the per-token \mt{update\_actor} time over the last 10
steps decomposes as $F/T+c$ = 36.8 + 26.7 (BF16), 53.3 + 14.6 (\tw) and 54.1 + 15.4\ms{} (\mxf). The measured
per-token times (63.9, 68.4, 69.2\ms) rank BF16 fastest, while the per-token cost $c$ ranks it slowest; the
contaminating term $F/T$ is larger than $c$ and moves the other way. The decomposition reproduces the measured values
to within 0.5\ms{} per token.

\paragraph{Fitting.} We regress each stage's time on $T$ = \mt{global\_seqlen/mean}, which in these runs
($\mathrm{dp}=1$) is the number of prompt and response tokens per step, over steps 4--60. With several data-parallel
ranks this metric is divided by the data-parallel size, and when a step aggregates several sub-steps verl sums times
but averages token counts; both must be corrected before fitting. The fit needs spread in $T$: $T$ spans 2.1$\times$ in
setting A, whereas in setting B it varies by only $\pm$10\% and the fit is ill-conditioned ($R^2$ from 0.03 to 0.98
across stages), so for B we use per-token means. Restricting setting A to steps with $T$ between 420k and 470k raises
the relative standard error of $c$ from 2.6\% to 35\%.

\paragraph{Uncertainty of $m^\ast$.} Within each fit the intercept and the slope are strongly anti-correlated
(correlation $-0.98$, because the data lie far from $T=0$), so the standard error of
$m^\ast=(F_q-F_b)/\bigl(N_{\text{mb}}(c_b-c_q)\bigr)$ must include each fit's intercept--slope covariance; treating $F$
and $c$ as independent overstates it about four-fold for \mt{update\_actor}. We report the delta-method error with the
full covariance (the two arms are independent runs) and, as a check, percentile intervals from a bootstrap that
resamples steps within each run; for \mt{update\_actor} the two agree. The residuals' lag-1 autocorrelation is within
$\pm0.2$, so treating steps as independent is a reasonable approximation for the timing fits, unlike for the mismatch
series.

\paragraph{Invalid regimes.} With parameter and optimizer offloading enabled (our first two rounds), the fitted
intercepts were negative, which is physically impossible and shows that transfer time varied with memory pressure; the
same comparison repeated on two hosts differed by 15 points on \mt{update\_actor}. We discarded those rounds.
Setting D needs offloading to fit in memory; we compare its arms only by their means on one host, never through
fits. Turning
offloading off halved the step time at identical reward (Appendix~\ref{app:incidental}).

\paragraph{Hosts.} On the slower-CPU host the GPU stages were 1.3--1.8$\times$ slower and the CPU-only advantage
computation 2.5$\times$ slower, while the most GPU-bound stage, the reference pass, slowed least (1.33$\times$); both
hosts report the same NVLink topology. We therefore compare timings only within a host.

\paragraph{Diagnostics we do not use.} verl also reports a sequence-level $\chi^2$, $\mathbb{E}[(\prod_t\rho_t)^2]-1$.
Over responses of hundreds to thousands of tokens it is dominated by a handful of sequences: in setting A its mean and
median order the two FP8 arms in opposite directions, and in setting B clamping makes most steps read $\approx-1$. We
report no result based on it.

\section{Per-stage time per step}
\label{app:stagetime}
\begin{figure}[ht]
\centering
\includegraphics[width=\linewidth]{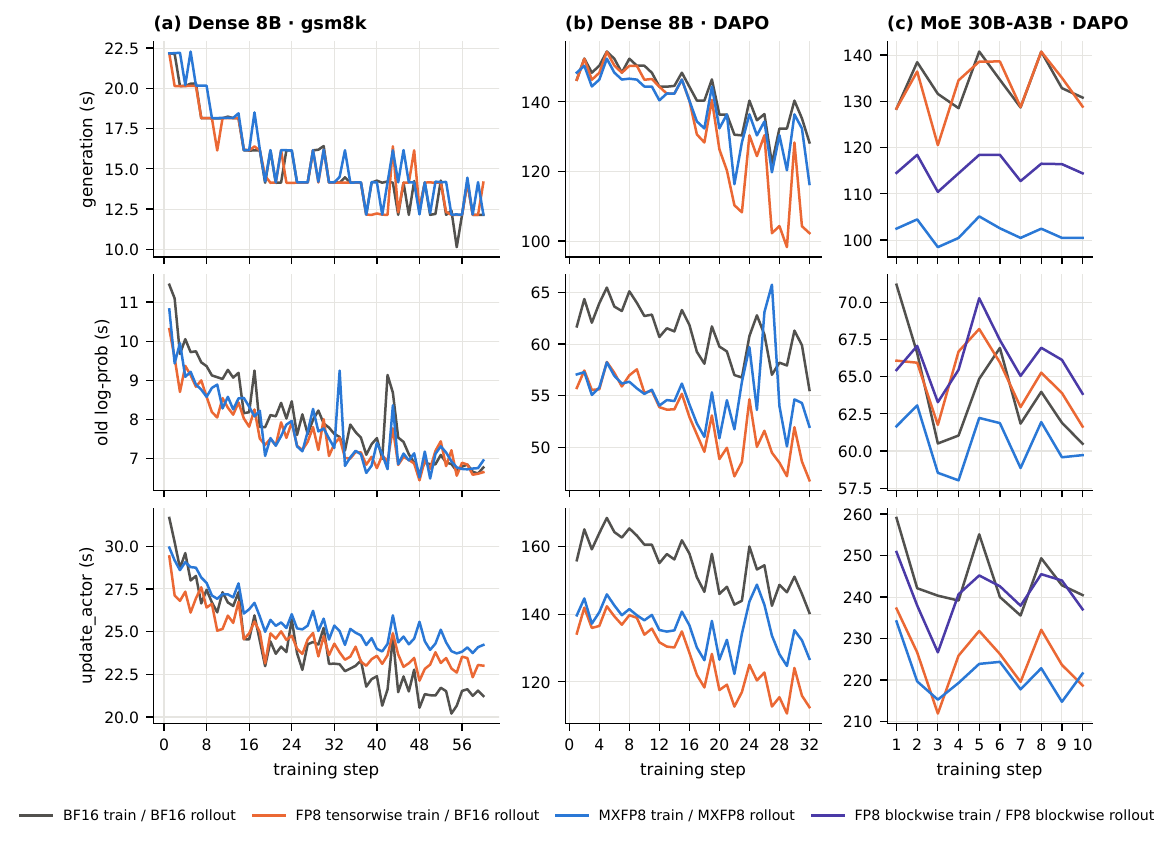}
\caption{Time per step of generation (top), the old-log-prob pass (middle) and \mt{update\_actor} (bottom), for the
arms and hosts of Figure~\ref{fig:steptime}.}
\label{fig:stagetime}
\end{figure}

Figure~\ref{fig:stagetime} splits the step time of Figure~\ref{fig:steptime} into its three largest stages, for the
same arms and hosts. Generation is unchanged by the recipes in setting A and nearly unchanged in B, where the
per-tensor arm's later steps are shorter because it generates fewer tokens. In setting D the \mxf{} sampler is the
fastest at every step, at about 100\,s against about 130\,s for the BF16 samplers, with the FP8-blockwise sampler in
between. The update stage of setting A shows the break-even point being crossed (Section~\ref{sec:breakeven}):
\mxf{} is faster in the first steps and slower once the micro-batches shrink.

\section{FP8 blockwise on Blackwell}
\label{app:blockwise}
On SM100, TE~2.16 converts each FP32 block scale to the E8M0 scale of an MXFP8 GEMM by keeping its exponent byte
(\mt{(sf >{}> 23) \& 0xFF}); both block-quantization kernels assert power-of-two scales on SM100, so a run with FP32
scales stops at the first quantization. On a B200 with power-of-two scales, \mt{te.Linear} under
\mt{Float8BlockScaling} gave relative errors against BF16 of 0.0377 (output), 0.0423 (dgrad) and 0.0462 (wgrad), and
\mt{te.GroupedLinear} 0.0376, 0.0475 and 0.0510, identical to four digits to \mt{MXFP8BlockScaling}. In a simulation of
the truncation that the assertion prevents, FP32 scales reduced to their exponent raise the GEMM's relative error to
0.53 (against 0.038 with power-of-two scales and 0.037 with exact FP32 scales).

\section{Incidental findings}
\label{app:incidental}
\begin{itemize}
\item \textbf{Offloading.} For Qwen3-8B on two B200s, disabling parameter and optimizer offloading reduced the step
time from 134.9\,s to 66.9\,s with identical reward (0.876), without running out of GPU memory.
\item \textbf{Inactive weight decay.} PyTorch's fused CPU AdamW applies decoupled weight decay as a multiplicative
factor $1-\eta\lambda$ computed in FP32. With $\eta=10^{-6}$ and $\lambda=0.01$ the factor $1-10^{-8}$ rounds to exactly
1, so weight decay had no effect in setting D, in all arms.
\item \textbf{Reference-model offset.} In setting D the reference model did not receive the actor's MoE configuration
(FP32 router, all-to-all dispatch, grouped GEMM), so the KL loss to the reference is nonzero at step~1 (0.0017 for
BF16, 0.0055 for \mxf, 0.0063 for FP8 blockwise) before any update. The train--rollout diagnostics compare the actor
with the sampler and are unaffected.
\end{itemize}

\section{Controlled experiments}
\label{app:controlled}

\paragraph{Configuration.} All 17 runs use setting A's command line (Table~\ref{tab:settings}; tensor parallelism
2 with sequence parallelism, verl's default) on two B200 GPUs of one host with an AMD EPYC~9555 CPU, and verl at the
commit of the original runs plus experiment-only hooks that time, log and dump data without changing the computation
(\mt{anc/\allowbreak controlled\_kit}); the synchronization timers do add a device synchronization at each of their
four phase boundaries, and time TE's quantization with CUDA events. The four matching runs use
learning rate 0 and FlashInfer autotuning and run 10 steps; the fully matched run runs 12, and its last two steps carry
the byte dumps. The partially matched sampler reproduces F2 by turning off verl's exact exclusion matching, so that
vLLM's substring fallback leaves \mt{gate\_up\_proj} in BF16 and each synchronization quantizes 180 of the 252 linear
weights instead of all of them. The sweep runs use learning rate $10^{-6}$ without autotuning and run 6 steps, of
which steps 2--6 are timed; the \mxf{} 4-sequence run runs 8, and its last two steps carry the byte dumps. A BF16
recording run (learning rate 0, 8 steps) stores every step's trajectories, which the sweep runs inject through verl's
rollout-skip mechanism in place of generation. The half-batch runs use 32 of each step's 64 prompt groups, chosen by
ranking the groups by token count and taking one of each adjacent pair, which keeps the length mix: the half batches
carry 0.4987 to 0.5009 of the full batches' tokens. Their mini-batches hold 16 prompts, so they too make two optimizer
updates per step. The two profiling runs repeat the 1-sequence size for 3 steps under
Nsight Systems~2025.6 (CUDA, NVTX, cuBLAS and UCX tracing, one report per profiled call), and we analyze step 3.

\paragraph{Statistics.} KL and the other diagnostics are ratios of sums over tokens or sequences, so per-prompt sums
over each prompt's 8 samples are sufficient; recomputed from them, the pooled KL of every step matches the logged
\mt{rollout\_corr/kl} to a relative $10^{-7}$. Because all arms see the same prompts at each step, the bootstrap draws
(step, prompt) units with replacement and applies the same draw to every arm, which removes prompt difficulty from the
differences. Steps 2--10 are the primary window, as for speed; including step~1 changes each KL reduction by at most
0.5 points (16.3\% for full matching against the BF16 sampler). In the sweep, each learner and micro-batch size
contributes five timed steps, paired across runs by step because all runs train on the same data. The reference pass, which is BF16 with the same micro-batch in every run, takes between 9.78 and 11.06\,s across
the full-batch runs; this run-to-run offset, which the fits do not model, is one reason the standard errors of $a$ are
large. Time outside every stage timer averages 0.03--0.08\,s per step in the sweep runs and about 0.2\,s in the
matching runs, whose diagnostic dumps fall outside the timers.

\paragraph{Byte comparison.} At the selected synchronization the hooks capture the engine's canonical weight and
scale after the last bucket is loaded and before vLLM's post-processing, and the post-processed tensors after it; on
the learner they capture the TE module's cached FP8 forward weight at the start of the next update and the BF16 weight
at the end of the update. An offline script quantizes the source weights with TE (the checkpoint at learning rate 0,
and the tensors received at that synchronization in the sweep run), maps Megatron's fused and sharded layout to
vLLM's (qkv rows regrouped by key-value group, TE's scale padding cropped), and compares bytes. The comparisons
are: A, the engine's canonical bytes against TE; D, the kernel layouts against the canonical bytes (the weight's
storage, and the scales against the cuBLAS 128$\times$4 swizzle of the canonical scales); B, the learner's cached
weight against TE; C, the learner's BF16 weight against the source; and E, the learner's weight before an update
against the one after the previous update. For freshness, a second script compares the engine's canonical bytes at
the two selected synchronizations, and the learner's BF16 and cached FP8 weights at the same two points. Every FP8 byte
that changed belongs to a BF16 element that changed, and a change of $d$ BF16 steps altered the E4M3 byte in a fraction
$\min(d,16)/16$ of cases, within 0.1 points for every $d$, as expected because E4M3 keeps 3 of BF16's 7 mantissa bits.

\paragraph{Profile.} Kernels are grouped by name: GEMM (cuBLAS and nvjet, including the MXFP8 ones), \mxf{}
quantization (\mt{quantize\_mxfp8}), scale rearrangement (\mt{swizzle}), NCCL, attention, normalization, and other.
For the timeline split the traces are exported to SQLite. Within the call's NVTX range, the intervals of kernels,
memory copies and memsets on all streams are merged into GPU busy time. Each remaining gap ends when the next GPU
operation starts, and the CUDA API call that issued it (matched by correlation id) splits the gap into time before
the call, when the host had not yet issued the operation, and time from the call to the start; a gap whose operation
was issued before the gap began counts as waiting on other work. For the not-yet-issued time we record whether the
issuing thread was inside a CUDA API call, a TE (\mt{nvte\_*}) or NCCL range, or none. A generic TE
quantization kernel writes the GEMM's scale layout when its template argument \mt{WITH\_GEMM\_SWIZZLED\_SCALES},
visible in the name, is true; the specialized cast kernel writes compact scales. TE~2.16.1 rearranges into a temporary
buffer per GEMM and keeps cached weights compact, so each micro-batch rearranges every weight's scales for its forward
GEMM and, in the update, its data-gradient GEMM: 144 of the 216 forward-only and 288 of the 576 update rearrangements;
the rest match the tensors that sequence parallelism all-gathers. Host time in TE's calls sums their NVTX ranges.

\section{Reproducibility}
\label{app:repro}
Each arm of a setting differs from the BF16 arm only in the precision flags:
\begin{itemize}
\item actor: \mt{fp8="e4m3"} and \mt{fp8\_recipe} $\in$ \{\mt{tensorwise}, \mt{blockwise}, \mt{mxfp8}\} under
\mt{override\_transformer\_config};
\item optimizer (blockwise arm only): \mt{fp8\_recipe="blockwise"};
\item sampler: \mt{rollout.quantization} $\in$ \{\mt{mxfp8}, \mt{fp8}\} for quantized samplers.
\end{itemize}
The ancillary files contain \mt{REPRODUCING.md}, with the command lines of all four settings as resolved at run
time, the exported per-step metrics of all runs, and the scripts that regenerate every table and figure
(\mt{scripts/\allowbreak analysis.py}, \mt{scripts/\allowbreak make\_figures.py}). The
controlled runs (Appendix~\ref{app:controlled}) are reproduced with \mt{controlled\_kit} on verl at the original
commit, and \mt{scripts/\allowbreak controlled.py} recomputes their numbers from \mt{data/\allowbreak controlled}.

\end{document}